\documentclass[conference]{IEEEtran}
\usepackage{cite}
\usepackage[pdftex]{graphicx}
\usepackage[cmex10]{amsmath}
\usepackage{amssymb}

\usepackage{times}
\usepackage{array}
\usepackage[tight,footnotesize]{subfigure}
\usepackage[font=footnotesize]{subfig}
\usepackage{graphicx}
\usepackage{mathtools}
\usepackage{microtype}
\usepackage{amsmath}

\usepackage{alltt}
\usepackage{shadethm}
\newshadetheorem{comment}{Comment}

\def\boxx{{\vcenter{\vbox{\hrule height.3pt
          \hbox{\vrule width.3pt height6pt
          \kern6pt\vrule width.3pt}\hrule height.3pt}}\;}}

\def\impos{{\;\vcenter{\hbox{\rule{5mm}{0.2mm}}} \vcenter{\hbox{\rule{1.5mm}{1.5mm}}} \;}}

\def\lrarrow{\leftrightarrow \kern-8pt \rightarrow}

\def\2{\frac{1}{2}}

\def\beq{\begin{eqnarray}}
\def\eeq{\end{eqnarray}}
\def\2{\frac{1}{2}}

\def\lrarrow{\leftrightarrow \kern-8pt \rightarrow}

\def\frightarrow{\rightarrow \kern-11pt /~~}
\def\reducesto{\simeq \kern -3pt >}

\begin{document}
\newcommand{\strust}[1]{\stackrel{\tau:#1}{\longrightarrow}}
\newcommand{\trust}[1]{\stackrel{#1}{{\rm\bf ~Trusts~}}}
\newcommand{\promise}[1]{\xrightarrow{#1}}
\newcommand{\revpromise}[1]{\xleftarrow{#1} }
\newcommand{\assoc}[1]{{\xrightharpoondown{#1}} }
\newcommand{\rassoc}[1]{{\xleftharpoondown{#1}} }
\newcommand{\imposition}[1]{\stackrel{#1}{\impos}}
\newcommand{\scopepromise}[2]{\xrightarrow[#2]{#1}}
\newcommand{\handshake}[1]{\xleftrightarrow{#1} \kern-8pt \xrightarrow{} }
\newcommand{\cpromise}[1]{\stackrel{#1}{\frightarrow}}
\newcommand{\policy}{\stackrel{P}{\equiv}}
\newcommand{\field}[1]{\mathbf{#1}}
\newcommand{\bundle}[1]{\stackrel{#1}{\Longrightarrow}}

\title{Testing the Quantitative Spacetime Hypothesis\\using Artificial Narrative Comprehension (I)\\~\\\normalsize Bootstrapping Meaning from Episodic Narrative viewed as a Feature Landscape}

\author{Mark Burgess}
\maketitle
\IEEEpeerreviewmaketitle

\renewcommand{\arraystretch}{1.2}

\begin{abstract}
  The problem of extracting important and meaningful parts of a
  sensory data stream, without prior training, is studied for symbolic
  sequences, by using textual narrative as a test case.  This is part
  of a larger study concerning the extraction of concepts from
  spacetime processes, and their knowledge representations within
  hybrid symbolic-learning `Artificial Intelligence'.

  Most approaches to text analysis make extensive use of the evolved
  human sense of language and semantics. In this work, streams are
  parsed without knowledge of semantics, using only measurable
  patterns (size and time) within the changing stream of symbols---as
  an event `landscape'.  This is a form of interferometry. Using
  lightweight procedures that can be run in just a few seconds on a
  single CPU, this work studies the validity of the Semantic Spacetime
  Hypothesis, for the extraction of concepts as process {\em
    invariants}. This `semantic preprocessor' may then act as a
  front-end for more sophisticated long-term graph-based learning
  techniques.  The results suggest that what we consider important and
  interesting about sensory experience is not solely based on higher
  reasoning, but on simple spacetime process cues, and this may be how
  cognitive processing is bootstrapped in the beginning.

\end{abstract}

\hyphenation{before}
\hyphenation{immun-ology}

{\small 
\tableofcontents
}


\section{Introduction} 

What might an eye see that has not seen before?  What might an ear
hear, never having heard a sound?  These simple-minded questions
underpin a significant amount of scientific and technological
discourse, and perhaps remarkably there is still plenty to be said on
such a fundamental matter.  

The Spacetime Hypothesis for concept formation\cite{cognitive}
proposes that patterns and meanings must initially be bootstrapped
from the simplest changes and patterns present in an agent's
environment. Those are changes observed in space and time---the only
source of information available to a cognitive agent {\em a priori}.
This work concerns how concepts may be identified from data streams,
how they are learned, and then how they are refined from the most
primordial observations. It is presented in two parts: this first
paper concerns the analysis of information arriving in realtime, based
only on a rudimentary discrimination of events in space and time. 
The second paper will consider whether meaningful narratives can be
generated from learned concepts based on spacetime relationships.

\subsection{Linguistics versus signal processing}

The setting for the work is narrative language, but not from a
linguistic perspective.  The study of language phenomena using
computational methods is an extensive field, dealing mostly with
statistical properties of grammatical language from large corpora of
text\cite{languagepatterns,feldman1}. The goal is typically to study
evidence of universal phenomena, positioning language as a coherent
and singular process. In technology, the goal of large data studies is
to find predictive models, inferring intent by matching patterns of
text or sight and sound to a dimensionally reduced set of outcomes
with simpler semantics. This is often used for command recognition or
biometrics.  Such a broad manifesto is fascinating, but more ambitious
than necessary here. The comprehension of limited forms of
communication, such as individual stories or narrative episodes, is a
different problem. Phrases within a confined storyline may be more
focused around particular subsets of language, essentially by their
more restricted intent. Thus, while the importance of rare phrases is
greater, the hope of seeing statistically significant repetition of
every common phrase is less reliable due to the limited context and
smaller amount of data. This is an interesting challenge.

Restricted narratives are interesting precisely because they are episodic,
time-ordered and experiential streams of data, which represent
the way we experience the world from a cognitive standpoint. Many of
those special effects are wiped out by averaging over large corpora from every
avenue of literature. They can be captured and summarized using a variety of
automated techniques, most of which rely heavily on semantic
modelling\cite{hovy1,textsummary1}. They are not limited to human
language either. Computer protocol exchanges, bioinformatic processes,
and all forms of process monitoring are basically narratives, often
with rather simple languages.  It's interesting to ask what common
principles explain the scaling properties of stories over these cases.

There is also interest on a philosophical level.  If one imagines the
emergent origins of cognition and language, at a time before humans
had their current skills, the signals which could be considered
meaningful would not be linguistic, but would have to stem from
simpler representations of changes those observers could detect around
them.  Those distinctions could later be represented as phrases of
language.  When corpus studies effectively look for echoes of those
processes, they `cheat' in the sense that they exploit our existing
present-day evolved understanding of language structure, using parts
of speech, word types, and even their semantics.  A primitive organism
or bio-mechanical process has no such rich inheritance to rely on to
decode communications. This is also the situation we are in when
monitoring technology and processes in the human-tech world.

From an agent perspective, linguistic communication is only as a set
of processes in which agents express `states of mind', according to
the processes that engage them.  Those ideas may eventually overlap,
from agent to agent, helped by communication channels and shared
media\cite{spacetime1,spacetime2,spacetime3}.  Communications then
range from initially primitive protocols up to extensive storytelling
with complicated grammars. In this work, the source of data will
natural language, mainly because this is readily available, but 
it will not be treated as language for the analysis, only as a stream
of changes.

\subsection{Communication without semantics}

Rather than jumping straight into the present evolved state of
language, with types, semantics, and even semiotics, the natural place
to begin is to look at text more like a time-series. This harks back
to the early work of Shannon in analyzing language as a stream of
characters in his now famous {\em Theory of
  Communication}\cite{shannon1,shannon3}. Shannon studied streams of
$n$-grams, or clusters of alphabetic characters, $n$ at a time, to
look for correlations and information content. On that level, it's not
possible to discuss meaning except in the broadest terms of scale.
Shannon's work has been elaborated upon by many authors since
(see the works of Montemurro and Zanette, for instance
\cite{montemurro3,montemurro1,montemurro2}).

The longer a sample of language, the more likely we are to be able to
identify statistically probable patterns within it, on the scale of
its smallest constituents. However, statistics have a basic flaw with
respect to quantitative measurement: they are based on scale invariant
ratios and do not naturally distinguish phenomena, they only count
populations of pre-known things.  Probabilities are not simply related
to the natural scales of a process (extent and duration), measured in
units of the observer.  To discover spacetime process variations, we
need to find a way around this.

The famous probabilistic studies of language made by Zipf and
Mandelbrot\cite{zipf1,mandelbrot1} revealed power law distributions
which have since become famous, but also less universal than
originally thought. From a spacetime perspective, a stream of
information is essentially a process trajectory. It has time base
(clocked by the arrival of alphabetic symbols) and it has extent (time
and space).  It would be preferable to have a model in terms of its
engineering dimensions, as one would expect in physics, and from there
identify coordinate invariants as significant phenomena.

Language studies tend to look directly at semantics, assuming a
pre-known lexicon. However, communication may vary in style and
substance, so the extent to which semantics are reliably invariant,
and intended meaning is conveyed, are subject to all kinds of
conditions. Meaning may be contextual, subject to errors, with
variable interpretations, etc.  Elaborate techniques have been used to
trick algorithms into `smart behaviours' at the expense of large data
processing resources\cite{neuraltext2,neuraltext1,word2vec}, with very
impressive results, but here the goal is to expend as little effort as
possible, from the perspective of a realtime observer, and consider
the principles by which such an agent might extract meaning from a
stream of data using only metric variations.

In short, unlike other studies, the goal here is not
to predict `what comes next' in a semantic stream, based on vast experience,
but rather to look for repeated patterns as {\em spacetime
  invariants}, within the process of communication itself.  This has
more in common with Shannon's early work than with modern Natural
Language Processing, and its goal is more limited than general
language comprehension: to find dynamical features in a stream that
may correspond to meaningful concepts. It relies on `shapes' rather
than on learned semantics.  As such the method is not limited to text:
it may also have implications for other `cognitive processes' such as
sequence analysis in bioinformatics and the monitoring of mechanical
and information systems.

\subsection{A parsimonious model of meaning}

Language evolves much like genetics, though mutation and selection
over a shorter timescale\cite{cavalli2000genes}.  Like all
computational processes, the budding linguist would have no lexicon to
start, and could only rely on simple discriminations of phenomena
observable in space and time. For instance, how events appear on
different scales, how to measure where changes start and stop, and so
forth.  This simple-minded idea of bootstrapping forms the basis of a
the `semantic spacetime'
hypothesis\cite{spacetime1,spacetime2,spacetime3}. That challenge has not gone
away---it persists, both where individuals seek to learn new
languages, and when we study domain-specific narratives where usage is
both specialized and unfamiliar. In many ways, the challenge is also
similar to what an immune system faces in learning about threats from
unexpected `phrases' of molecular genetic material, which it seeks to categorize
and respond to, often without the benefit of generational
learning\cite{forrest1,forrest2,forrest5}. I'll comment more on this
below.

As a model of these ideas, one can examine the {\em small data} in ad
hoc narrative. Then, instead of thinking of language as an established
system of types (nouns, verbs, etc), one may consider fragments of
language more like populations of `molecules' in a chemical
interaction---but in which one doesn't know the chemistry a priori.
Text is scanned using a sliding window to sample substrings of a
stream of text, building on Shannon's idea of $n$-grams across several
hierarchical scales.  Some words and phrases are ubiquitous, as if
other more specific fragments were dissolved in a solution of them.
Those common parts might catalyze an interior binding of phrases but
contribute little to the exterior function of the sentence.  Using this
general approach one may treat a data stream as a feature-scaling
chemistry problem to be unravelled.

\begin{figure}[ht]
\begin{center}
\includegraphics[width=7.5cm]{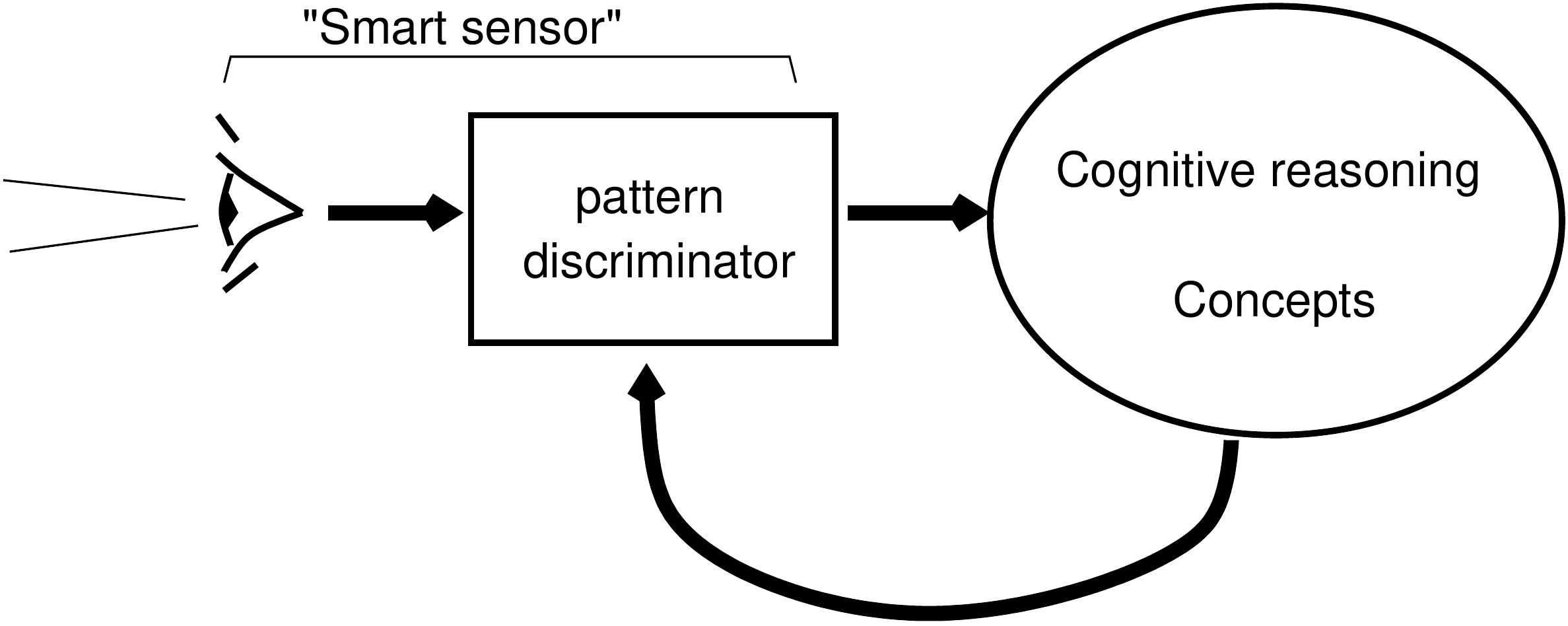}
\caption{\small A smart sensor is the first stage of a cognitive
  system, consisting of an entry port for data and specialized
  pre-processing, based on experience accumulated over long
  timescales.  Its learned (evolutionary) adaptations facilitate the
  realtime processing of streams for continuous cognition.\label{cognitive}}
\end{center}
\end{figure}

\subsection{Spacetime interferometry versus immunology}

The approach here is to look for {\em discriminators}, which might form a
basis set of sensory signals, against which to compare data streams.
The aim is to discover and evolve this set of key elements for efficient
recognition by a trained sensor. Candidate patterns can be
learned in realtime, and compared by a simple kind of interferometry
using short-term memory, even as a data stream unfolds in realtime.
This is in contrast to attempting a {\em post hoc} analysis based on a
frozen corpus.

Realtime interferometry is widely used in physics.  The idea is to
overlay sequences, side by side, and combine them in order to increase
the resolution and distinguishability of features which overlap in the
sample sets, amplifying the smallest differences by using one version
as the partial calibrator of another.  This combination of methods
deliberately avoids the prior accumulation of `big data', and instead
looks for a more realtime approach based on recent process history.
In that sense, it is similar to Bayesian methods\cite{pearl1}.  It can
still accumulate learning over large datasets, but now we can also
study the bootstrapping of meaning from dynamical processes as we go.
In particular, by focusing on only small amounts of data, realtime
context can be identified from causal parameters in the immediate
environment of an agent as data are received, thus providing
additional constraints on the meaning from a natural scope. This
means we are not diluting information about context and specialization.
Once basic data discriminators have done their work, one may then go
on to apply a promise theoretic approach to extracting a `smart
relational network'\cite{cognitive} (this will be considered in
a follow up paper), but we first need some empirical basis for discriminating
data samples.  

Information Theory recognizes data as a series of symbols belonging to
a fixed alphabet.  The challenge with scanning unknown data is to
first learn that alphabet, based on generic criteria over multiple
scales: character, word, sentence, paragraph, etc. The alphabet may
not be what we think it is. Specifically, one finds that the smallest
meaningfully distinguishable units of information are not the ASCII
character set, but rather word fragments.  This is where the
bioinformatic or immune system connection enters: short and long term
learning fragments of chemical symbology is essentially what an immune
system does for antigen epitopes---and in
realtime\cite{perelson1,perelson2,forrest1,lisa98283}).  Immunology
uses the the promises of patterns (here represented as ASCII symbols
and their recombination into larger effective symbols) to compare
their information directly. Changing concentrations of patterns in
blood are less important than the identification of particular
sequences considered
harmful\cite{perelson2,matzinger1,aickelin02danger}.  Only later,
might there be a link between population and significance, owing to
the process that manufactures a response. In bioinformatics, genes are repeated
many times over samples, and they are both relatively few in number
and all on the same scale, so one can use statistics in a predictable
way. In narrative, patterns have a much larger alphabet (i.e. lexicon)
and the possible combinations are very large, leading to very little
repetition over the lifetime of a narrative.  In order to think of
narrative in the same way as genetics, one would need vast corpora,
which would average over the interesting characteristics.  The way
meaning is extracted is an altogether different strategy.

Immune systems eventually tolerate certain patterns as harmless or
`self' in the same way that animals get used to their own smells until
something changes. Changes are easy to find on a single scale There
are multiscale phenomena in bioinformatics too: proteins could be
viewed as multiscale narratives, formed from processes derived from
genetic sequences.  The avoidance of false positive detection in the
immune system is believed to be mitigated by the requirement of
multiple confirmatory signals. Only the `co-activation' of certain
signals in the same context (spacetime frame) leads to a response.  We
can explore whether there are similar co-activation patterns in text
narrative for meaningful parts. One possibility might be the use of
`emotional assessments' combined with specific symbol sequences.  

The weakness of immune systems is that they can only look on a single
scale of molecular epitopes or genes, which are expected to be
repeated often on the scale of an organism or population, so one ends
up looking for outliers from a fixed or slowly varying set.
However, in narratives there is an explosion of new combinations of characters
and words from the outset, even in more limited language use. In order
to turn pattern discrimination into a reliable `smart sensor'
methodology, we need to find the limits on significance of this
explosion of pattern. The hierarchy of features is almost certain to
play a role in scaling this: from a small set of characters one sees
words, from the words come phrases, and from there sentence events.
Rather than approach this by brute force, the alternative is to study
shorter examples of narrative using principles based on an cooperative
structure, and try to identify when emergent patterns stabilize based
only on spacetime process discriminators.

In narrative, text is rarely repeated verbatim, except for small
phrase fragments, so how could statistics work? Narrative has an
intended direction.  It doesn't meander around random inputs, except
at a local level. There is a progression, so it's more akin to a
landscape of features than a soup of similar antigen fragments.
Nevertheless, a method of atomization (like the fractionation used in
organic chemistry) can still be used to find important scales, by
superposing the same stream at shifted intervals and looking for
interference, i.e. which phrases reinforce and which do not.
In the spacetime approach, we do expect to understand the fragments of
text that are extracted---only determine those to which meaning may
naturally be attached due to their structural importance\footnote{In a
  sense, meaning is defined as the opposite of entropy, with attention
  to scaling.}.  Their meaning might then be evaluated
by a human arbiter.

On larger perspective, this methodology also effectively comments on an
important and fundamental tension between probabilistic methods and
the measurement of dimensional scales (in the sense of the physical
dimensions mass, length, time, etc) in empirical science.
Probabilities are dimensionless scaling ratios, scale invariant and
not easily distinguished in multiscale phenomena. They relegate scale
to classification ranges that have to be fixed from the start.  This
makes sense when the ensembles over which probabilities are computed
are fixed and unchanging. But when scales themselves are changing it
can induce a blindness to pattern. In a spacetime approach, one can
only compare repeating processes in order to calibrate one against
another.  That is not a statistical process, but an ongoing
adaptation. The superposition (or parallel execution and comparison of
trajectories) of processes substitute for statistical populations by
reinforcement with alignment of trajectories rather than stationary
probabilities.  The prevalence of statistical techniques
notwithstanding, we cannot model real world phenomena in order to
extract semantics without proper and dynamical identifications of
scale.

\subsection{Study overview and structure}

The structure of the paper is as follows: we begin by looking for what
regular and invariant features can be measured from individual
narrative texts, using interferometry to look for the basic scales and
phenomena in a stream of text. A series of experiments, performed on
`small data' (of the order of kilobytes to megabytes) probe the
effectiveness and limits of the method.  This serves also to establish
contact with other work, and determines key scales of the problem. The
role of these episodic experiences in finding principles behind a
`smart sensor' is the larger goal.  We want to know what properties
hold for smaller samples, so we look at documents one by one, but test
the assumptions across many different cases.

For convenience, English documents have been used exclusively, though
from widely different sources and purposes. The specifics of English
are not used in any way, except for what can be discerned
statistically as patterns.  Treating sentences as `events' arriving at
a receiver, we then use knowledge of scales to look for a spacetime
definition of importance in patterns of text. From those definitions,
we can then extract the most important sentences in order, and see
whether they capture the meaning of the narrative, without
understanding the language.

The question is then: how shall we assess whether phrases extracted in
this way are representative or not of the whole narrative? This turns
out to be a non-trivial question.  We have a choice: (i) to adopt a
kind of Turing Test approach to naturalness, i.e.  if a human assesses
the result as `satisfying' the this is a step in the right direction,
with our already advanced understanding of semantics, or (ii) to look
for invariant principles or `rules' that we discover by learning or
some other theoretical means.
The first challenge is to not be biased by one's own understanding of
documents---merely trying to represent and reproduce a human
understanding of the data. What we want to do, instead, is to look for
principles based on spacetime pattern discrimination and used these to
see what kinds of stories emerge from the stacking of concepts
according to the statistical invariants, metric and graphical
principles described in \cite{cognitive}.  As always, with a
non-trivial idea, this is a lengthy business in which one tries to
avoid fooling oneself with smoke and mirrors. 

An interesting side-effect of this to study the usefulness of the
Semantic Spacetime Hypothesis, which has an interesting overlap with
the techniques of the Immunity Model\cite{burgessC11,burgessC14}:
would scaling principles for an emergent `semantic chemistry' suffice
to make a reasonable attempt at story comprehension and generation?
The ultimate goal of the study is to take an input stream and turn it
into a reasoning system in the form of relational promise
graph\cite{promisebook}, as described in\cite{cognitive}. The
link between narrative and reasoning was initially discussed in joint
work with A. Couch in \cite{inferences,stories,cognitive,spacetime3}.

\section{The statistical mechanics of stories} 

\subsection{Statistical properties of stories within language}

Documents in natural language can be found everywhere. On the WWW, we
have free access to mostly short snippets of language, like online
articles, blog-posts.  Occasionally, we find entire books online that
can be studied\footnote{Especially useful is the corpus of older and
  classic books at \cite{gutenberg}.}.  By the standards of machine
learning, even a lengthy book is `small data'---hardly representative
of the entirety of its mother language.  But the goal here is not to
study an entire language, only stories by various authors.  

Some obvious questions arise. Could one find certain fingerprints in
the use of language by a particular author, or over the changing
epochs of history? The corresponding problem for network traffic has been the holy grail
for network intrusion detection for
years\cite{lisa971,barbara1,zanero1,stolfo1}. Should we view grammar
as a set of rules or as something to be learned from emergent pattern?
Classical approaches to language using Hidden Markov Models and
parsing trees have attempted to view language as a logical exercise,
based on regular semantics, but more recently the successes of machine
learning in language recognition have shown how stochastic pattern
matching is a more likely explanation for the emergent structure of
language.

Studies of language, by restricted corpus, are discussed often
in linguistics\cite{languagepatterns}. As a `type
system' language reveals certain patterns classifying identifiable
`parts of speech' \cite{statbasisforlang,statbasisforlang2}. These
conventional identifications might not be useful
for classifying documents according to the constraints of narrative.
Sometimes too much attention to semantic `data type' only prevents other natural
identifications from being made.
Corpus studies of language look for meanings in phrases and
grammatical nuggets.  For example, in \cite{languagepatterns} we find
many interesting aspects of language in general but---as often the
case in linguistics---the arguments are based on logical analyses, or
rule-based grammatical thinking.  The presumed importance of logic and
thus grammar is thus one of the more inexplicable aspects of classical
linguistics---there seems to be little evidence of a logical
structure; there is more evidence of a relational structure (as with other forms
of network association).  Type basis
assumes cognitive faculties to discriminate fixed types and roles within
language from the outset, without a natural bootstrapping mechanism.
Such things may become universal in nature, or the domain of only
humans after millennia of learning---yet the origins of such
discriminations are what one would like to discover independently of
prior assumptions \cite{statbasisforlang,statbasisforlang2}.  This
study offers some possible answers based on very simple arguments,
without reference to presumed semantics or universal grammars.

Stories have a stronger intentional focus than random linguistic interaction.
As we read a report, or listen to a story, there is an unfolding progression of
concepts, which gives the listener a unique train of contextual cues.
The paradox of general language comprehension is that, the more
repetition one accumulates (meaning that they become more conducive to
statistical analysis), the less significant these cues become.
Resolving that paradox is an essential part of realtime cognition---and it's what
distinguishes realtime cognition from large data trained machine learning, which
mimics the process of evolutionary adaptation more closely than it does cognitive adaptation.

\subsection{Data sources}

For data, a number of documents of different lengths were used with
deliberate contrast (see table \ref{tab1}).  The Gutenberg Project is
an excellent source of documents, available in standard formats. As
author, I also picked a number of self-written sources for which I have the
complete text in a readable format. In each case, initial preambles
with copyright messages and contents lists etc were stripped off
manually as these are not part of the narrative and serve only to
confuse the meaning.  In a more sophisticated semantics-based analysis
they could be distinguished as `types', but that goes beyond the
present scope.

\begin{table}[ht]
\begin{center}
\begin{tabular}{|c|c|c|l|}
\hline
Characters & Words & Paras & Name\\
\hline
1050785 &  179511 & 3739 & History of Bede\\
1206540 &  208458 & 2540 & Moby Dick\\
1150632 & 192106  & 961 & The Origin of Species (6th)\\
45686   &    8113 & 138 & Random sentences\\
1729903 &  261132 & 685 & Slogans in HTML\\
1278574 &  224738 & 8290 & Slogans in \LaTeX\\
6918    &    1038 & 16 & Test 1 HTML (partial doc)\\
141795  &   20267 & 277 & Test 2 HTML (full doc)\\
81953   &   12812 & 165 & Test 3 HTML\\
39530   &    6412 & 135 & Test 4 HTML (partial doc)\\
312137  &   51279 & 795 & Test 5 HTML (full doc)\\
\hline
\end{tabular}
\bigskip

\caption{\small A few example data samples used, showing length in characters and words.
The subject matter covers a wide range, from technical documents to history and even fiction.\label{tab1}}
\end{center}
\end{table}

The examples in table \ref{tab1} show a sample of documents examined.
Their subject matter varied from technical papers to blog posts to
monographs and novels.  The document `Slogans'\cite{slogans}, which is
a novel is of comparable length to the historical work of Bede, has a
very different character and style and therefore provides a contrast
to emphasize the non-universality of narrative: in other words, a
distinction that is not based on the amount of data.

An immediately apparent difference between the data concerns the way
that paragraphs are counted. Far fewer paragraphs are used and counted in the
novel, rendered in HTML format, but far more paragraphs are counted in
the \LaTeX version. This is an artifact of the surplus of character
dialogue in a novel, where each new piece of dialogue between
characters generally starts a new paragraph. However, the scanning
algorithm could choose whether to see this as a new paragraph of not.
The choice is arbitrary, but it clearly affects the `measure' of the
document. Clearly, we cannot avoid recognizing key signposts offered
by punctuation. So, if a smart sensor is to be based on unbiased data
from a stream, it needs to be able to distinguish measures by whatever
incoming patterns are available.  Clearly, recognizing a few
punctuation marks is not as drastic as recognizing `parts of speech',
like verbs and nouns, etc, as one does in grammatical analysis.  These
kinds of issues eventually have to be turned into a universal policy.

Perhaps the simplest resolution to this issue is to assume that the receiver can only promise
a finite size `buffer' for parsing sentences (related to its short
term memory).  I refer to this buffer size as a `leg' in the journey
of the narrative.  A size of about 200 sentences (two orders of
magnitude) was settled on for a decent compression rate in stream
capture, though tests showed that the results were not strongly
dependent on this scale, as it was overshadowed by the definition of
`meaning' (see section \ref{meaning}).

\begin{figure}[ht]
\begin{center}
\includegraphics[width=7.5cm]{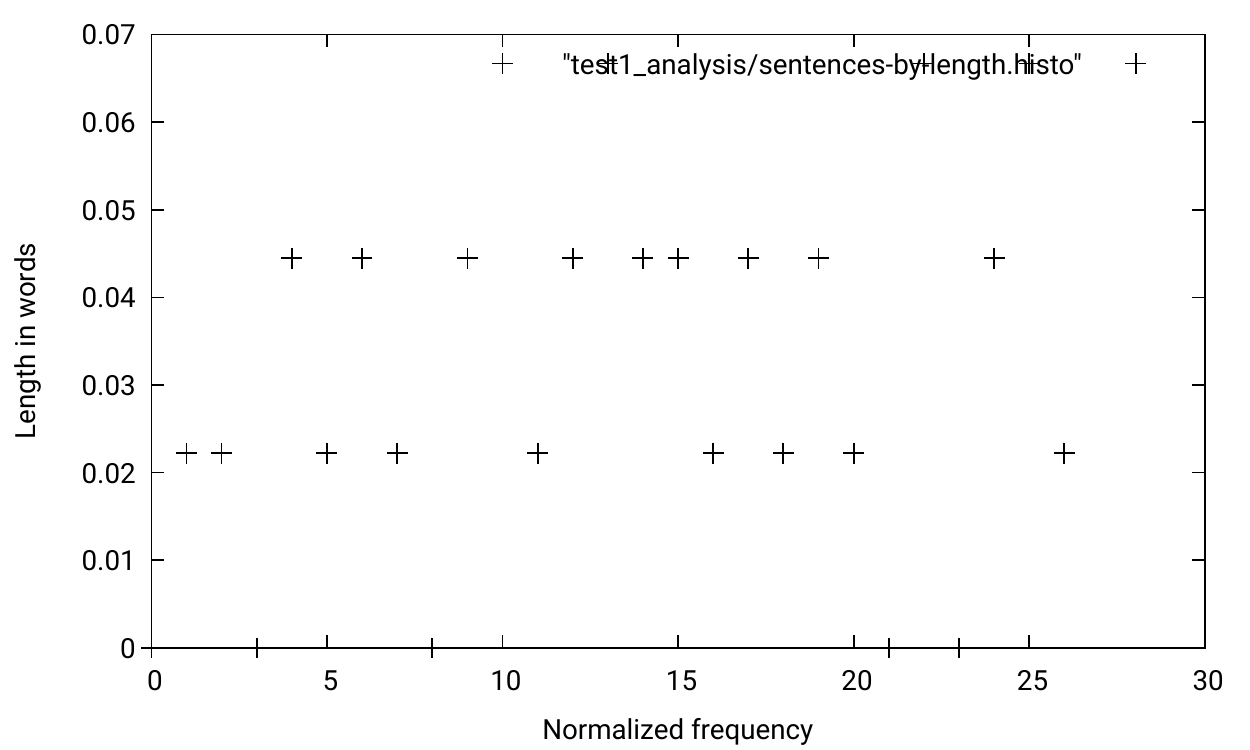}
\caption{\small Profile of normalized frequency (probability) of
  finding sentences of a certain length in a short blog post sample of
  around 1000 words. It's plain to see that the document is much too
  small to generate a stable distribution of sentence sizes---and yet
  it's still possible to handle documents of this size using spacetime
  discriminators.\label{sentence_test1}}
\end{center}
\end{figure}

\begin{figure}[ht]
\begin{center}
\includegraphics[width=7.5cm]{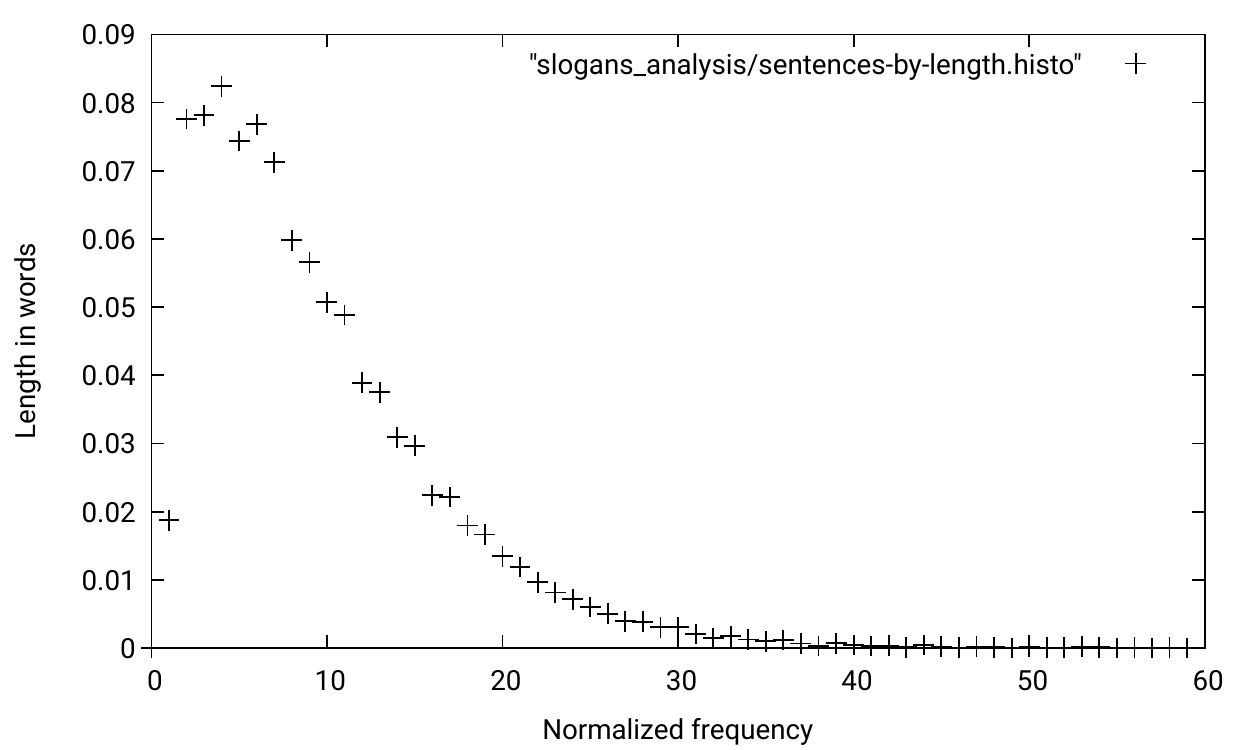}
\caption{\small Profile of normalized frequency (probability) of
  finding sentences of certain lengths in the Slogans novel of 180,000 words.
The profile is different from figure \ref{sentence_bede}, for a document
of comparable size, indicating that this could be used to distinguish documents
to some extent.\label{sentence_slogans}}
\end{center}
\end{figure}

\begin{figure}[ht]
\begin{center}
\includegraphics[width=7.5cm]{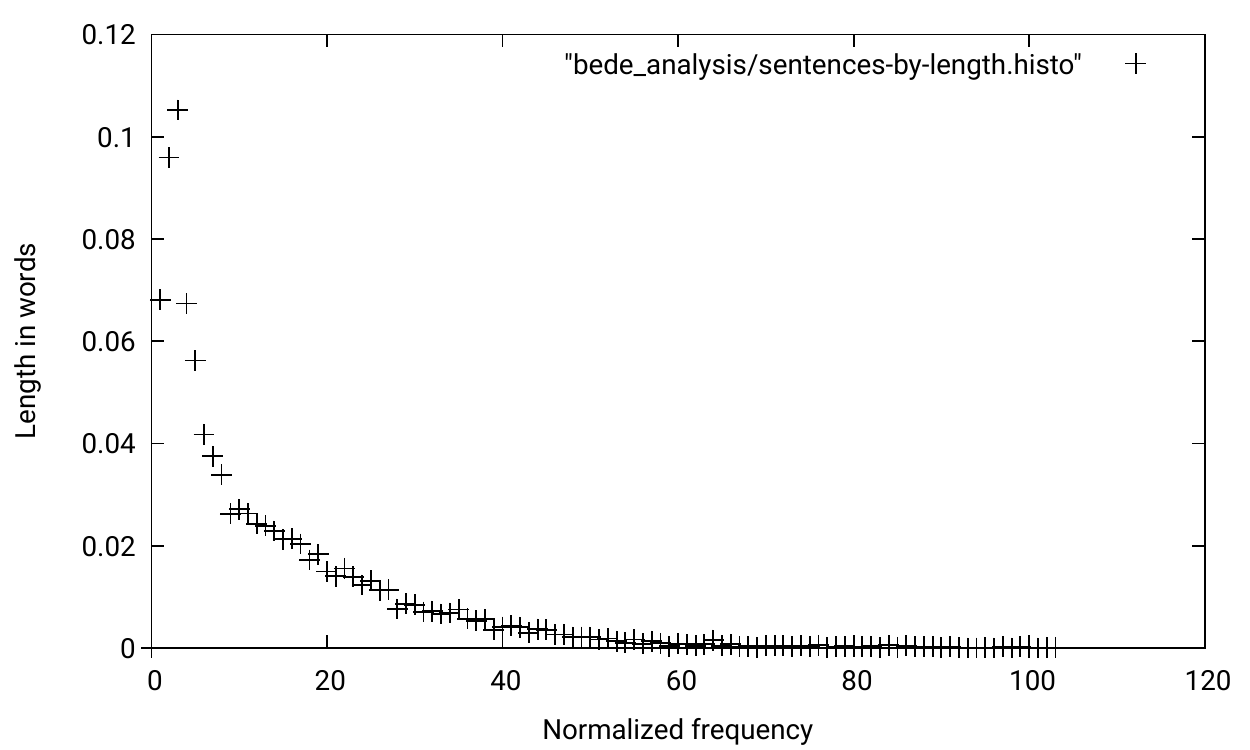}
\caption{\small Profile of normalized frequency (probability) of
  finding sentences of certain lengths in Bede of 130,000 words.
The profile is different from figure \ref{sentence_slogans}, for a document
of comparable size, indicating that this could be used to distinguish documents
to some extent.\label{sentence_bede}}
\end{center}
\end{figure}

So the measure of time, within a narrative, is related to each agent's
own sampling rate and memory capacity rather than what the source
tries to impose upon it. This feels like a more realistic (as well as
less arbitrary) resolution, as it is dictated by the boundary
conditions of the system, which is always an important (if not
underrated) source of scale information.

\subsection{Document dissociation}

To analyze narratives, and find the durable scales of interest, I use
an approach that follows the basic methodology of Promise Theory,
first looking for agents and then their promises. Thus, we first
atomize a system into its smallest functional parts\cite{promisebook}.
The process is analogous to what one does in chemistry to analyze
similar molecules, e.g. DNA. By breaking into small parts and looking
for differences and repeated patterns, one can identify a `table of
elements' for the important functional pieces, and compare what they
promise in the context of the experiment. By further using `constructive
interference' between samples, trying all possible phase alignments along a
process timeline, one can effectively set up an interferometry problem to single out
the significant elements and reduce noise.

The space of patterns in documents (called epitopes in immunology)
consists of generalized alphabets.  Narrative documents have obvious
structures: a character set, words, sentences, paragraphs, chapters,
and so on.  The promises of words are not significantly different from
those of single characters.  Indeed, they are on approximately the
same scale; some words are represented as only a single letter. Words
extend the character sets of alphabetic languages by recombination,
and in mnemonic languages, such as Chinese, words and characters are
basically indistinguishable.

\begin{figure}[ht]
\begin{center}
\includegraphics[width=7.5cm]{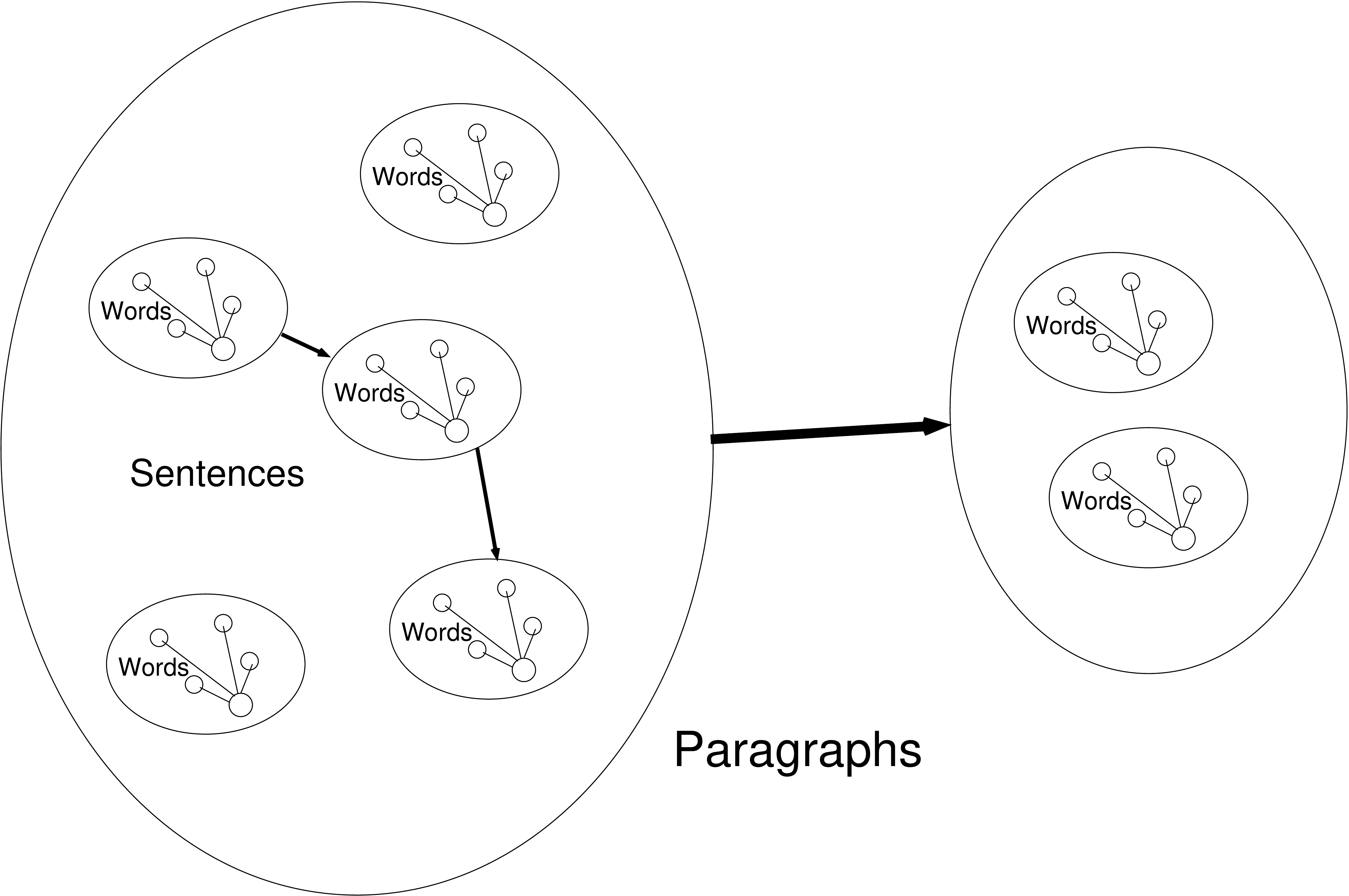}
\caption{\small A document has major structures like words, sentences, paragraphs. The order of
words within sentences is not strongly defined; the order of sentences within paragraphs
may convey an order of events, and similarly for paragraphs. There is a differences
in roles between large scale and small scale cooperative structures.\label{docstructure}}
\end{center}
\end{figure}

We can take words as the smallest agents of language, and look at how
patterns of words form invariant or repeated clusters (superagents).
Clusters, which we can call phrases, may be of any length.  Sentences
provide natural boundaries for meaning.  A cluster of $n$ words is an
$n$-phrase (denoted $\phi_n$). If we sample a document sentence by sentence we could expect
to find repeated clusters of words up to the length of a sentence.
Phrases do not span multiple sentences, as a matter of definition, but
one could find repeated patterns on any scale in principle.

Within each sentence, word order is quite flexible -- words don't
often promise specific orders, nor do sentences.  However, some words
promise not to end a phrase (e.g. "but", "and", "the", "or", "a",
"an"), because they bind to words strongly so that finding such words
at the end of a fragment would imply its incompleteness, just as a
free radical in chemistry would signify an incomplete molecule.
We may or may not wish to make use of these kinds of promises to make
a smart sensor.

Sentences promise collective
meaning, so they seem to act as superagent clusters of words in a
promise theoretic sense. We can try to test this by looking at
examples of larger data samples.  Within a paragraph, the order of
sentences is partially important: it might explain causal
order---though this is a simplistic approach to narrative. Modern and postmodern
narratives often play with storytelling order for artistic effect. 
For simplicity, we can ignore this issue for the present work, and assume that
sentences assume a partial ordering within paragraphs. In a similar way,
we may assume that paragraphs are partially ordered to reflect causal
order.

To decompose a document into identifiable fragments, we could adopt
any number of approaches. A `Millikan experiment' or fractionation
approach (such as one might use in organic chemistry) to looking for
meaningful atoms is adopted here. There is no unique way of doing
this, so it's worth comparing a couple of significant alternatives. We
can try to break up sentences into random pieces and look for the
smallest invariant pieces. Initially, confining ourselves to English
texts, it's not too unconventional to assume that the smallest
meaningful fragment is a word\footnote{Shannon showed that this was the 
primary unit of information\cite{shannon1}.}.  Longer fragments (compound words, noun
phrases, adjectival phrases etc) also behave as larger units, all of
which must be shorter than a sentence.  Two approaches are contrasted
here:
\begin{itemize}
\item {\em Unbiased pattern population sampling}: Sliding windows or
  `round-robin' buffers of different word lengths move across
  sentences picking all possible combinations. We count these to see
  if they are ever repeated. If so, they have some invariant meaning.
  Those which are not repeated are treated as random. This might
  occasionally be false for a single narrative, viewed across language
  as a whole, but for the documents core meaning it's a reasonable and
  unbiased assumption. Given the average length of sentences in the
  documents studies was of the order 10-20 words, phrases of up to 6
  words were looked retained in the analyses here.

\item {\em Local metric changes in spacetime landscape}: The most common patterns---those words
  which distribute homogeneously like salt and pepper throughout
  documents (e.g.  `of' and `by', etc), contrast with the outstanding
  topographic features of any spacetime process. Given their repeated
  usage, the homogeneous parts might be considered relatively
  meaningless, even when they service functions, like covalent
  bindings (`can of worms' etc), so one could try to use them as if they were punctuation to
  split sentences into fragments---a bit like breaking up DNA into
  partial strands.
\end{itemize}

Phrases of different lengths can be expected to behave differently as
meaning accumulates words into repeated clusters.  Why don't 6 phrases
behave like 1 phrases on a larger scale? The answer must be that they
cannot bind as easily to other fragments around them, constrained by
the length of a sentence. These are measurable hypotheses that can be
tested using narrative data of different lengths.

There is clearly a temptation to rush into a single scale analysis, as
many authors do, using experience of empirical rules for language and
the knowledge of semantics obtained by studying larger language over
many years.  However, would be jumping the gun. Much of what's written
in specialized literature makes unique and technical use of language
in its narrative, something like a pidgin, so there's value to being
able to derive meaning from pattern in a fundamental way rather than
relying on the spoils of evolutionary selection. That's something we
can test.

Here, the goal is the apply principles of {\em invariant pattern}
identification over space and time to the recognition of stable `epitopes', or `atoms'
and `molecules' of meaning to which semantics would later be
attached\cite{cognitive,spacetime3}.  We could then dissociate these
phrases and try to find where meaning is localized in narratives. This
would be a first step to mapping out narrative with a metric
coordinate system. This, in turn, might later be used to order and
explain the structure of narrative for the purpose of explaining and
generating artificial patterns of reasoning.

\subsection{Statistical distributions and scaling features}

Without prior knowledge, it makes sense to begin with the simplest
kinds of measurement to chart out the approximate scales and measures
of the problem. Does one find an average sentence length, average
paragraph length, etc, within narrow limits? Or is there a long-tailed
distribution of these measures? If word order were absolute, phrases
would bind into strict conditional chains (like Markov processes), but
if word order is in fact free, then words would form sentences that behave as
network-like clusters rather than linear rule-based processes, which
in turn suggests power law structures\footnote{Exponents and power
  laws arise when what happens next depends on what's already there.
  Linear behaviour arises when what happens next is independent of
  what's already there, as in a Markov process. Exponents spell
  interior dependency.}.  This can be tested, and indeed the latter seems to
be case.

\begin{figure}[ht]
\begin{center}
\includegraphics[width=7.5cm]{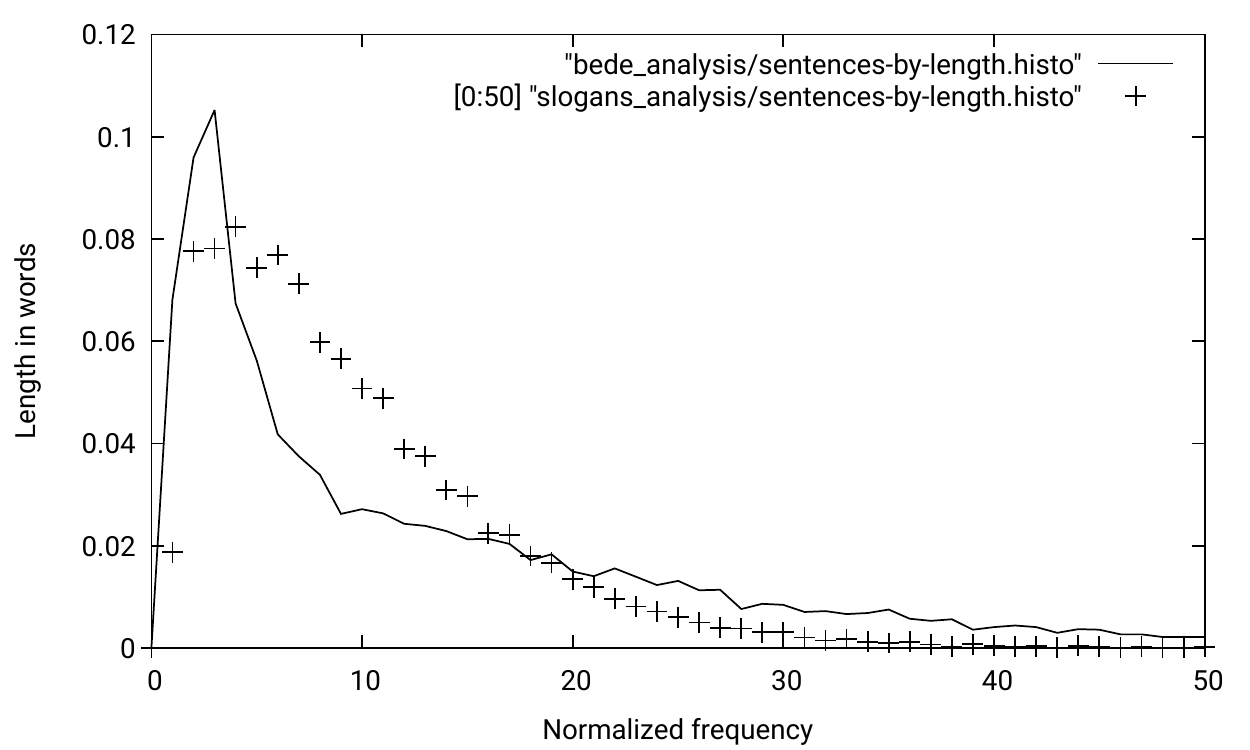}
\caption{\small Comparing profiles of similar documents by normalized frequency (probability) of
  finding sentences of a certain length, indicating that this could be used to distinguish documents
to some extent. The data from Bede is joined into a solid line to make reading easier.\label{sentence_comp}}
\end{center}
\end{figure}

Looking at the basic scale of sentence lengths, one does find some initial
convergence to a `stable distribution' over a document, at least by author style (see
figure \ref{sentence_comp}).  Paragraph lengths, on the other hand
remain more noisy, but one might expect these to converge too, given
longer coherent documents. There are orders of magnitude difference in
average sentence length (10-20) versus average paragraph length
(150-300) in various documents. This might be a result of personal
author style or of document style.  This reveals a technical problem,
however: encoded document representations are highly irregular. The
way we represent and parse documents leaves the detection of paragraph
boundaries quite difficult to standardize.  Some representations use
new-line breaks after each line, and double breaks for paragraphs,
e.g. poetry.
\begin{quote}
\tt
Mark had a little lamb\\
Whose fleece was white as snow
\end{quote}
The same text might be written;
\begin{quote}
\begin{center}\tt
Mark had a little lamb whose\\ 
fleece was white as snow and\\
everywhere that Mary went hi\\
s Lamb was sure to follow wi\\
thout a pensioner's bus pass\\
\end{center}
\end{quote}
Paragraphs are technically harder to identify, parse, count, and interpret, because they are
used differently in various kinds of document. For instance, in a
novel each time a person speaks modern authors tend to start a new
paragraph.  
\begin{quote}
\tt
Then it all began to unravel:\\
``You're dumb!''\\
``No, you're dumb!''\\
It was pure chaos.
\end{quote}
But is this part of one paragraph, or one for each voice? Should we
count this kind of speech as a special case? To do so would be to
resort to semantic distinctions of the kind that this study attempts
to ignore. A better solution is to disregard paragraphs altogether as
an unreliable measure. Promise theoretically, relying on what is given
by source or donor (as so-called ``+'' promises) is a misplaced
strategy anyway. Each agent observer (receiver or acceptor) is the
ultimate arbiter and bottleneck on observation, through its ``-''
promises, so we should adapt a comprehension based on receiver rather than
source.

These conventions will affect the counted length of scales normally
attributed to paragraphs--and it means that we can't easily compare
different forms by encoding without some rigid policies up
front\footnote{This illustrates a weakness of purely quantitative
  measurement in diverse settings: identifying appropriate invariants
  across semantically different domains is a precarious business---and
  the hopeful belief in a kind of Central Limit Theorem to wash away
  sinful semantics is misplaced.}. The resolution to this matter is to
forget about paragraphs and replace them by `legs', discussed below in section \ref{multiscale}.

\begin{figure}[ht]
\begin{center}
\includegraphics[width=7.5cm]{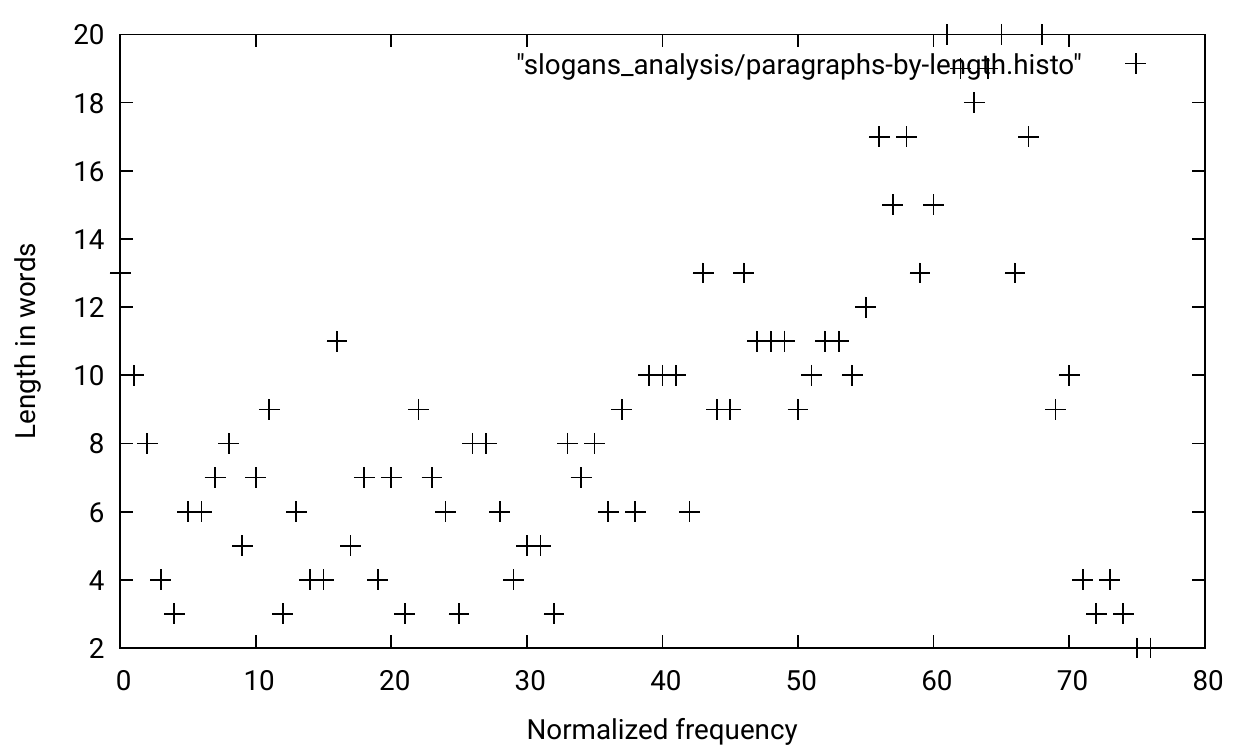}
\caption{\small Normalized frequency of paragraphs by increasing length (in words)
for novel Slogans. Despite being of comparable length to figure \ref{para_bede},
the pattern is erratic as the document counts far fewer paragraphs due to the 
technical formatting of dialogue, etc.\label{para_slogans}}
\end{center}
\end{figure}

\begin{figure}[ht]
\begin{center}
\includegraphics[width=7.5cm]{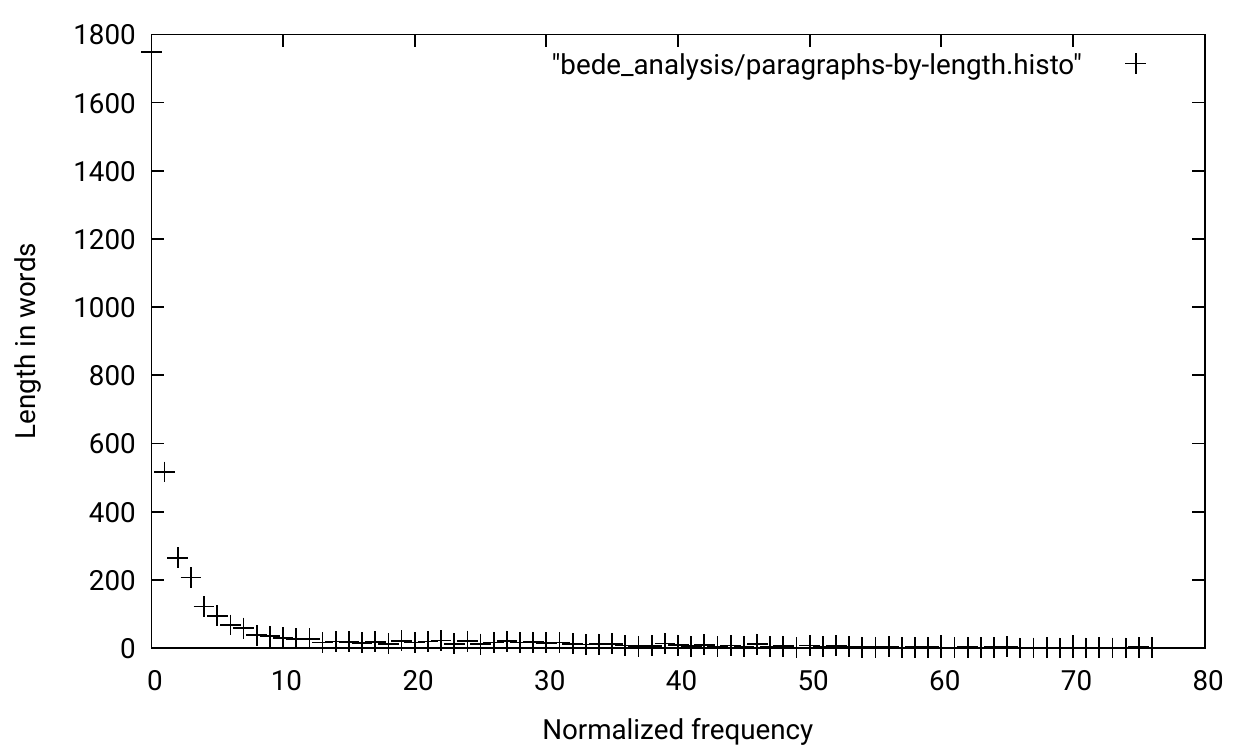}
\caption{\small Normalized frequency of paragraphs by increasing length (in words)
for Bede's history. The tighter distribution seems to follow from the more regular
counting of the document, without the kind of technical formatting used in a novel
like Slogans.\label{para_bede}}
\end{center}
\end{figure}

The number of long and short paragraphs in documents apparently
follows a rudimentary power law, as shown in figures \ref{para_slogans}
and \ref{para_bede}, no matter how we count them. There's a preponderance of shorter paragraphs
and long sentences are rare events. That this is true in a given style
is one thing---to see it across vastly different styles and formatting
types is rather more convincing. 

Early studies by Zipf\cite{zipf1} and Mandelbrot\cite{mandelbrot1}
also discovered power laws between word classes, and have been given a
lot of weight in network and complexity science, where power laws are
considered to be a sign of universality\cite{newmanreview}.  In
section \ref{powerlaw}, I'll show how this observation is somewhat
vacuous.  In \cite{zipflanguage}, the author explores the extent to
which statistics depend on meaning. Here, the implication of the
spacetime hypothesis is that this is upside-down: meaning derives from
frequency distribution. What is missing from ensemble statistics is
that we need to view distributions of words in space and time as
significant---not simply total frequency over a corpus.

\subsection{Multi-scale events}\label{multiscale}

Characters and words form an extended combinatoric alphabet---they are
part of a hierarchical lexicon of symbols that express single
patterns. They combine to into sentences which are the complete and
unique spacetime `events' in a stream.  There is no functional
difference between words and characters. Words simply play the role of
combinatorially extended characters, delimited by spaces---as one sees
from audible language, where they are actually representations of
phonemes, emphasizing the spacetime process aspect of language.

A sentence emerges as a single `event' in the sense of a dynamically
and semantically stochastic arrival. Sentences, rather than words
express complete statements. We define a sentence by the text between
full-stops (periods) `.'  or `?' or `!'.  Events do not distinguish
themselves strongly by interior promise or a `chemistry' of words and
types within, only by their exterior promised characteristics.
However, we can make note of all the fractions in the decomposition of
a sentence event for later analysis.  At this stage we are
making no assumptions above promised semantics---only spacetime
structural promises.

How we treat paragraphs is a different matter.  Although paragraphs
are a tool for writers to use in order to distinguish their intended
reading of breaks and changes, there is no obvious beginning or end of
a narrative on the scale of the story, except the start and end of the
document. The length of paragraphs depends on style. Any breaks are a convention by mutual agreement between
source and receiver, and shifts in topic are usually gradual and
enacted over multiple sentences or paragraphs.  This leaves the
sentence as the primary distinguishable event.

\subsection{Invariant fragments, sample size and repetition rate}\label{powerlaw}

\hyphenation{sentences}

To apply the Spacetime Hypothesis, we want to look for the natural
process structures, delimited by empty space, or by the most common
patterns.  Sentences are the natural choice, and `legs' are sampling
measures that limit perception. End of sentence punctuation and spaces
provides the most common markers, which suffice to discriminate
sentence boundaries.  Other common patterns (such as actual words)
could also be explored as possibilities for fractionation.  

\begin{figure}[ht]
\begin{center}
\includegraphics[width=7.5cm]{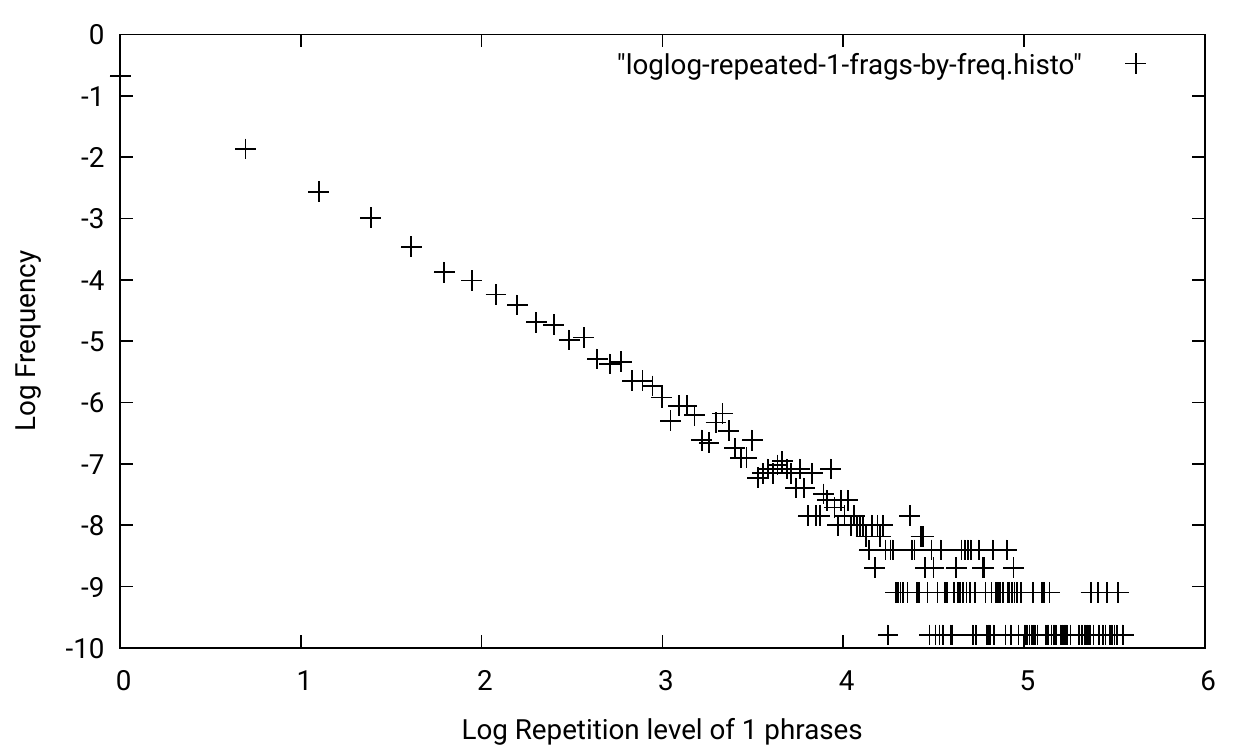}
\caption{\small Log-log plot of the meta histogram for repeated
  1-phrases. The horizontal axis shows the number of times certain
  phrases are repeated, the vertical axis shows the frequency. Most
  words are repeated a few times; a tiny number are repeated many
  times (such as `the', `of', `by', etc).\label{repeat1}}
\end{center}
\end{figure}

\begin{figure}[ht]
\begin{center}
\includegraphics[width=7.5cm]{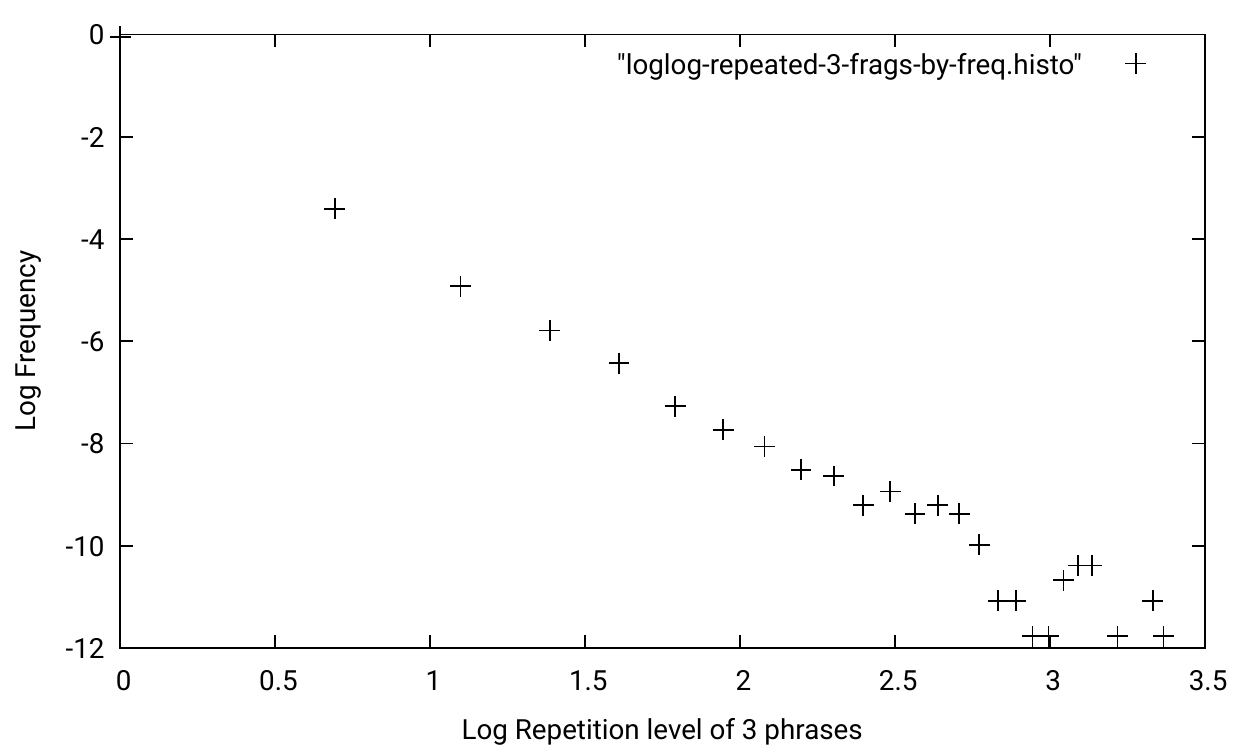}
\caption{\small Log-log plot of the meta histogram for repeated
  3-phrases. This shows the extension of binding behaviour from
figure \ref{repeat1}. By the time we reach 6-phrases there are too
few repetitions to form a plot.\label{repeat3}}
\end{center}
\end{figure}

Documents are the largest containers I'll consider here. They are
assumed to tell a single story, with a common subject. The sources
chosen are not anthologies or multi-subject compendia, to avoid adding
unnecessary noise.  Given the generality of the hypothesis, it's
perhaps unsurprising that there are general patterns that emerge for
sufficiently long documents.  Typical examples are shown in figures
\ref{repeat1} and \ref{repeat3} for 1-phrases (words) and 3-phrases
(e.g. `words in groups').

\begin{figure}[ht]
\begin{center}
\includegraphics[width=7.5cm]{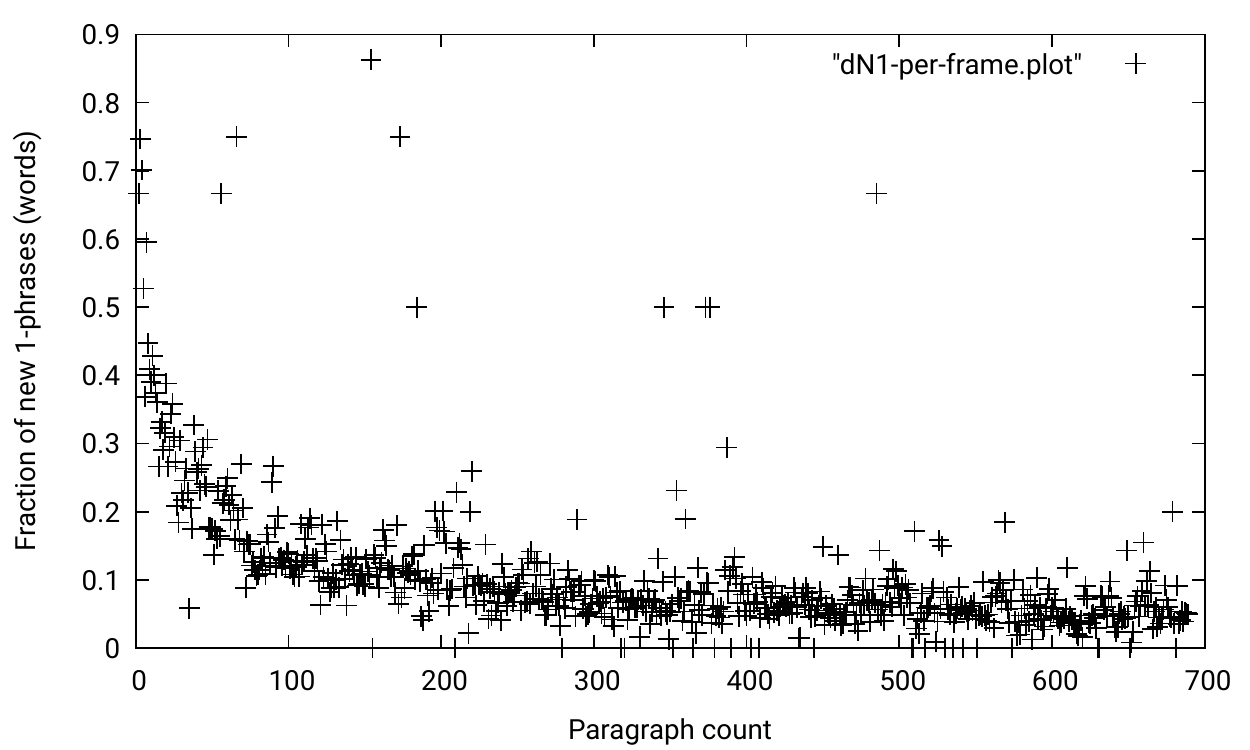}
\caption{\small From a 120,000 word document, on a consistent narrative, the rate of new words
per paragraph shows an exponential decay.\label{dN1}}
\end{center}
\end{figure}

\begin{figure}[ht]
\begin{center}
\includegraphics[width=7.5cm]{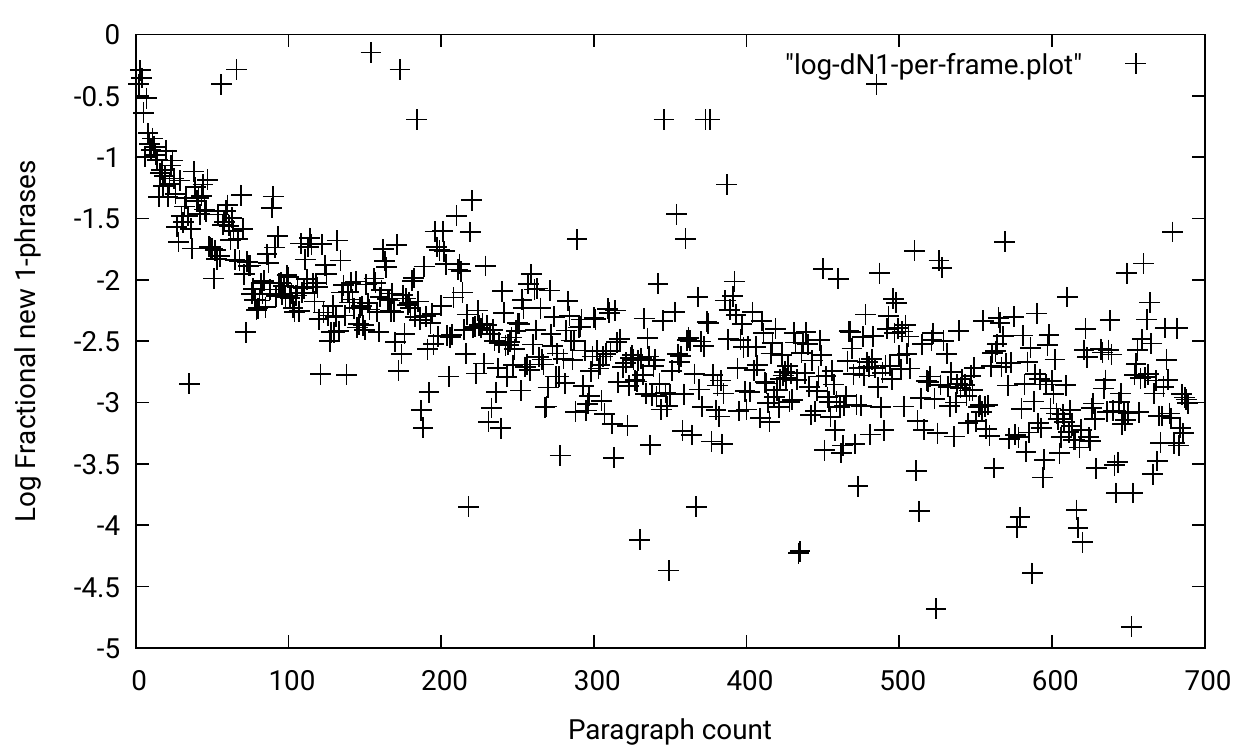}
\caption{\small The logarithm of figure \ref{dN1} which expects a straight line to test for exponential decay. We still see
some sign of decay, suggesting a power law behaviour (figure \ref{loglogdN1}).\label{logdN1}}
\end{center}
\end{figure}

\begin{figure}[ht]
\begin{center}
\includegraphics[width=7.5cm]{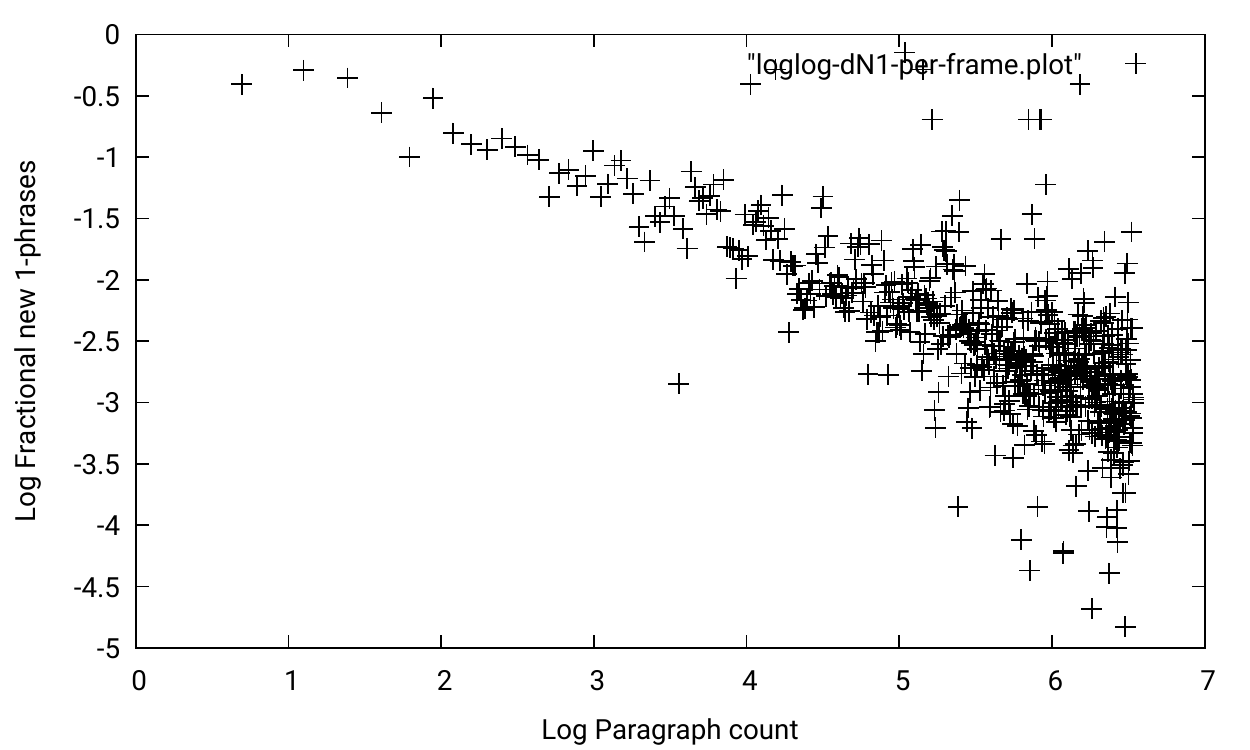}
\caption{\small The log-log version of figure \ref{dN1} to test for power law behaviour.
Apart from the increasing scatter over the `proper time' of the document, this is a convincing
straight line for small times, which becomes increasingly erratic at longer process times, especially
for lower counts, which possibly indicates the limits on the supply of new words. Each language
has a commonly-used core which will eventually disrupt the assumption of constant novelty. This pattern has
been observed widely in \cite{zipflanguage}.\label{loglogdN1}}
\end{center}
\end{figure}

\begin{figure}[ht]
\begin{center}
\includegraphics[width=7.5cm]{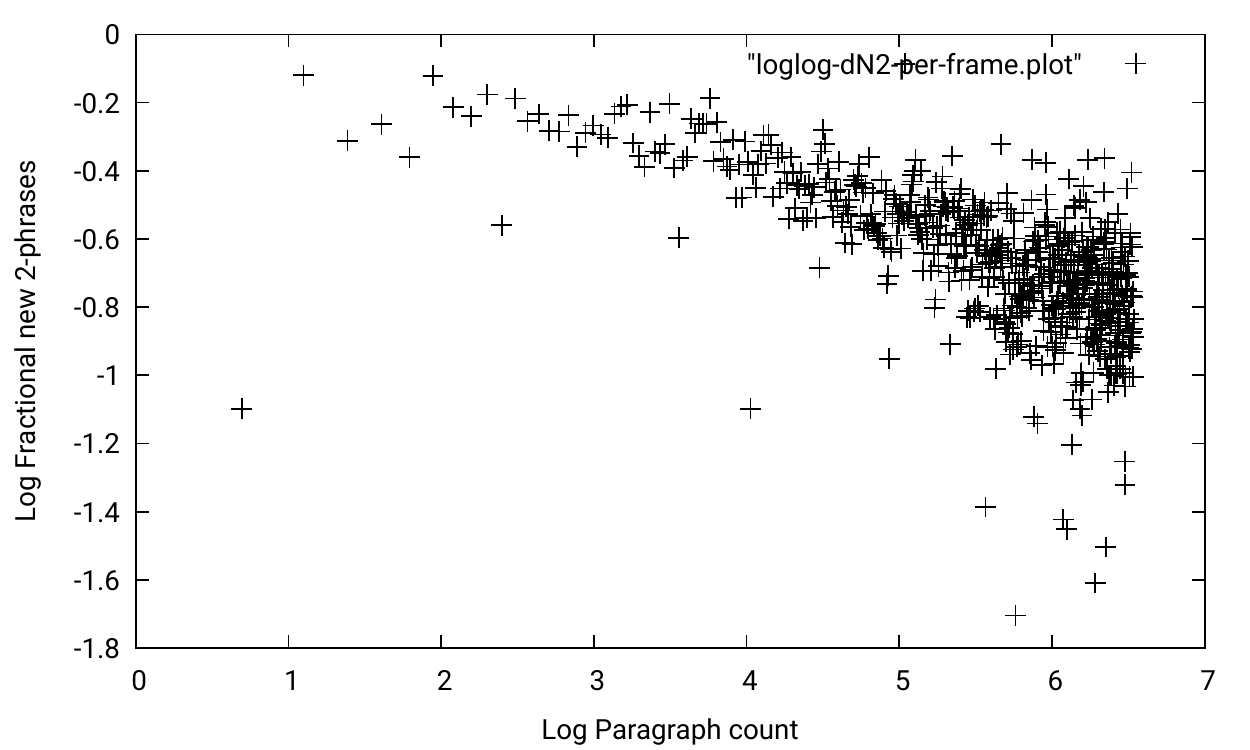}
\caption{\small The log-log plot of 2-phrases, showing similar power behaviour but weaker
than for single words.\label{loglogdN2}}
\end{center}
\end{figure}

\begin{figure}[ht]
\begin{center}
\includegraphics[width=7.5cm]{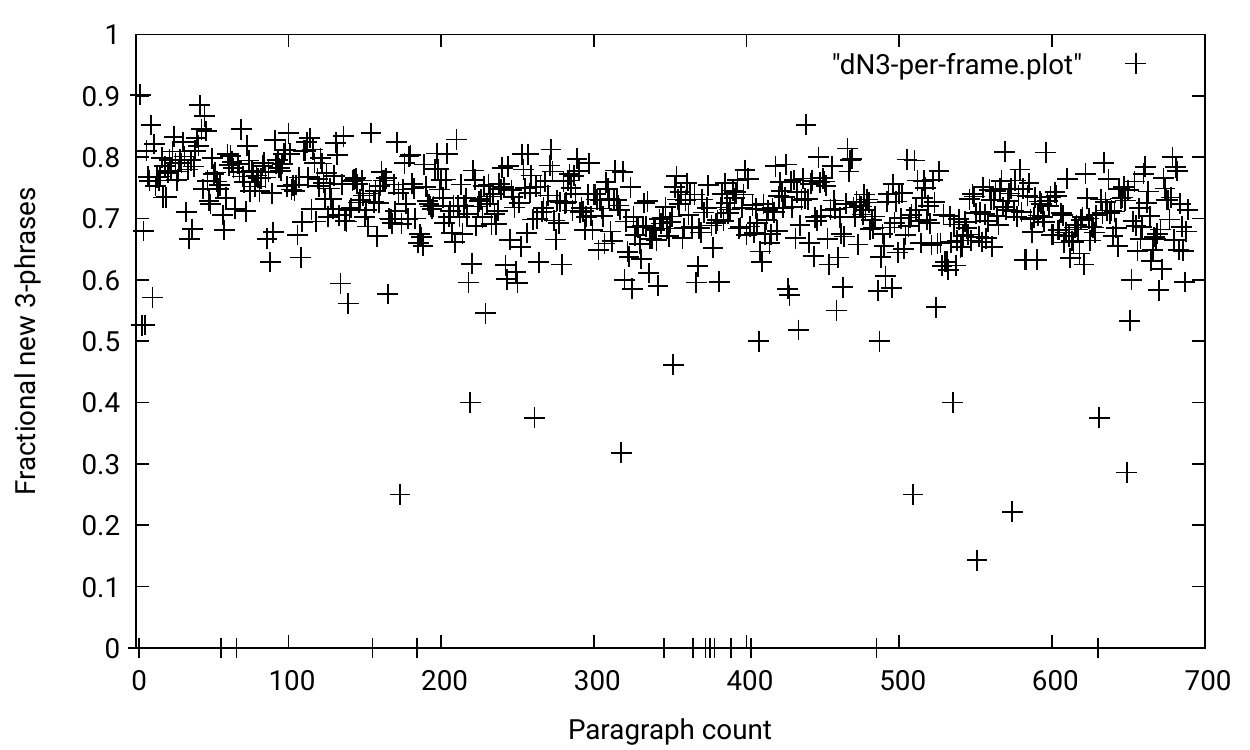}
\caption{\small The non-log plot for 3-phrases already shows a collapse of the power law
behaviour at 3-clusters and higher orders. From this size, the rate of new arrivals
is approximately constant, although widely variable.\label{dN3}}
\end{center}
\end{figure}

\begin{figure}[ht]
\begin{center}
\includegraphics[width=7.5cm]{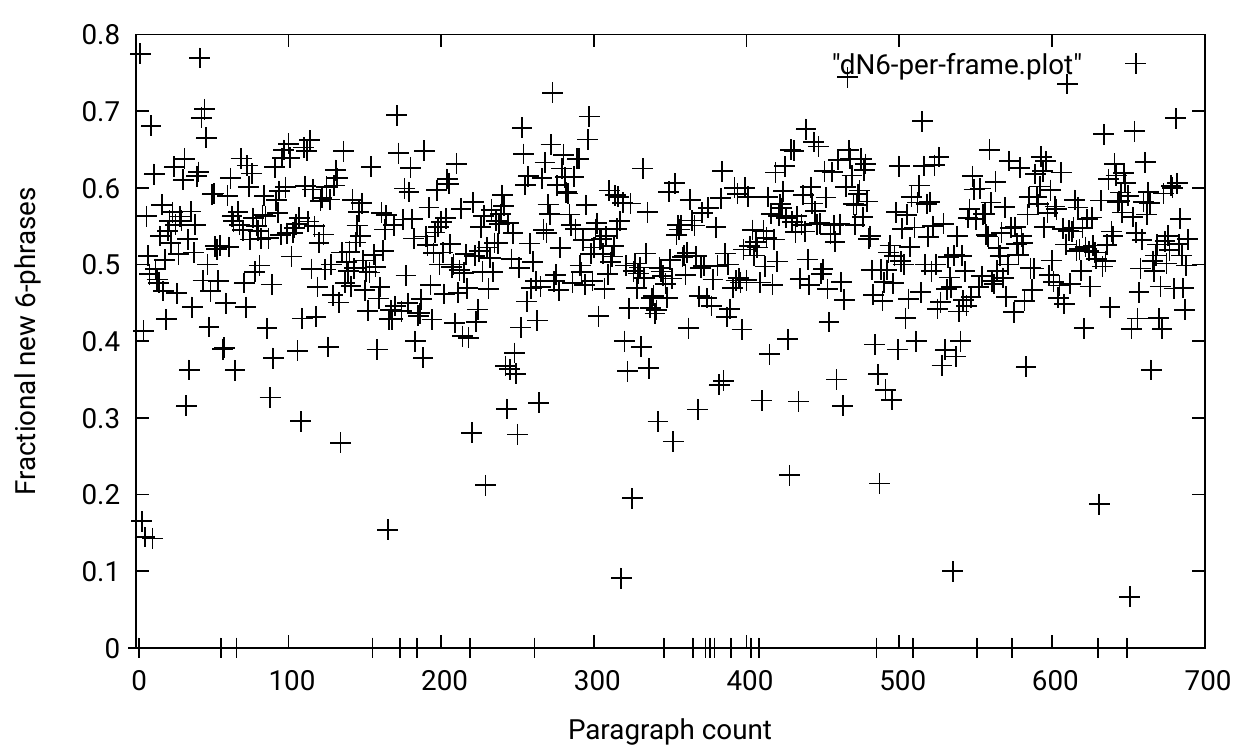}
\caption{\small The non-log plot for 6-phrases already shows a collapse of the power law
behaviour at 6-clusters and higher orders. From this size, the rate of new arrivals
is approximately constant, although widely variable.\label{dN6}}
\end{center}
\end{figure}

Let's define the number of distinct words already seen by $N$, and
$dN$ is a change in that number caused by the arrival of a word that
we haven't seen. To measure this number, any cognitive process has to
have some amount of memory. For any finite agent, the amount of that
memory is also finite.  Figures \ref{dN1}, \ref{logdN1},
\ref{loglogdN1}, \ref{loglogdN2}, \ref{dN3}, and
\ref{dN6} show a variety of plots of $n$-phrases of different length.

The predilection for `big data' statistics notwithstanding, we can make simple
predictions concerning what distributions of new
minimum size for statistical stability in a
sampling process. The samples used here are shown in table \ref{tab1}.
The assumption of a `smooth' (i.e. differentiable) approximation to
the process of learning should be quite good for longer documents,
and be a rough approximation to short ones.
If we have a finite amount of memory to remember $N_0$ things, then
arrivals will appear to be new at a decaying rate that depends on how much
we already know. The decay rate of
new words, assuming a homogeneous random process, would be a simple
exponential decay, where the rate of newness decays according to
the amount of memory already consumed:
\beq 
\frac{dN}{dt} &\propto& (N_0 - N)\nonumber\\
N(t) &\sim& e^{-\lambda t}
\eeq
if the decay were simply a matter of unbiased arrival and learning as we go,
over some timescale $\lambda$. Initial inspection of figure \ref{dN1} might seem
to confirm this view; however, the logarithmic plots in figures \ref{logdN1}
and \ref{loglogdN1} reveal that the plot is actually a power law.

For longer phrases ($n > 1$), which are rarely repeated due to their
increasing specificity, it would be reasonable to assume that the rate
of new phrases would be basically constant throughout a story---this is seen
as a constant flat rate in the non-logarithmic plots of figures \ref{dN3} and \ref{dN6}.  On the
other hand, if one chose only particular significant repeated
$n$-phrases (denoted $\phi_n$) then it would be expected to decay exponentially just like
single words (because they function like single words (e.g.  `son of a
gun' etc.).  The longer the phrase, the more likely the decay rate
will be zero on average.

Figure \ref{dN1} shows the arrival rate of new words over the course
of a document.  It looks superficially like the simple exponential
decay, described above. However, plotting logarithmically in figures \ref{logdN1} and
\ref{loglogdN1} reveals that it's actually a power law decay.  This
might be assumed to come from a model of preferential attachment, as
discussed in \cite{watts1,linked}.  Some words and phrases, after all,
function as space fillers and supplementary punctuation have network
binding roles to form effective network clusters. However, as I'll show
below, that observation is a somewhat vacuous artifact of the way
we choose to count---deliberately rendering measures scale invariant
in order to make a simpler comparison.
Functional binding words that form implicit network bindings between
the parts of a sentence tend to promise constant valency for binding.
In a linear stream, they can only bind to one or two other agents,
so there will be a cutoff in the interactivity of $n$-phrases.

The handwaving exponential argument above doesn't take proper account
of the quantity measured in the graphs however. This is an artifact of the desire to
eliminate spurious scales, for scientific comparison, by expressings fractions relative to the actual length
of sentences---but these are also the very features that we want to use here.
In this sense, the appearance of power laws is misleading and somewhat vacuous: an artifact
of the construction of using probability rather than metric scale to represent the system.
Figures \ref{dN1}-\ref{loglogdN1} count the number of previously
unseen words in a number of sentences $s$, measured in words ($\phi_1$).
The number of such new words over the interval $ds$ is $dN(\phi_1)/ds$,
but the plot is also normalized by the length $s$ which is proportional to $s$
in stochastic irregular units. Shifting to dimensionless ratios, we can
use the probability of seeing a new $\phi_n$ which is the probability that
we have not already seen it before $\text{Pr}(\neg \phi_n)$. Now the number
of new arrivals over a sample $s$ is $s \text{Pr}(\neg \phi_n)$, so 
\beq
\frac{dN}{ds} = s \frac{d\text{Pr}(\neg\phi_n)}{ds}
\eeq
For consistency, this must be proportional to the probability that we haven't seen the
fragment before (thanks to our choice of a dimensionless ratio), 
which by definition is $\text{Pr}(\neg\phi_n)$. One ends up with an almost vacuous
statement about coarse-graining. Thus, we have a self-consistency equation for the
probability:
\beq
 s \frac{d\text{Pr}(\neg\phi_n)}{ds} = -\beta(n) \text{Pr}(\neg\phi_n),
\eeq
for some $\beta$ which is independent of $s$, but which may depend on the fragment length $n$.
This simple `renormalization group equation' may be integrated to yield the general form of
the probability which is a scaling law for figures \ref{dN1}-\ref{loglogdN1}:
\beq
\text{Pr}(\neg\phi_n) \propto \left(\frac{s_0}{s}\right)^\beta,
\eeq
where $s_0$ is a constant scale, representing an average sentence `leg',
and $\beta$ is a positive constant which determines the decay rate.
This gives a formally infinite measure for zero time ($s = 0$) but that
is excluded in practice. 

When no $n$-phrases are ever repeated, then $\text{Pr}(\neg\phi_n) = 1$, so
$\beta(n) \rightarrow 0$. From the promise bindings of the smallest words, 
we would expect $\beta \rightarrow 0$ at about $n=3$, since this is the longest
fragment that could bind independent of sentence context (which is a separable
scale than the word chemistry, analogous to what a chemical promises to do in a mixture
with rather than what bond affinities it promises). This suggests 
\beq
\beta (n) = \beta_0 \; \theta\left(\frac{n_0-n}{n_0}\right),
\eeq
where $n_0\simeq 3$ over the statistical scales\footnote{Fragments
  behave essentially like Dirac sea fermions.}, and $\theta(x)$ is
the Heaviside step function which is $1$ for $x > 0$, zero for $x < 0$, and $\2$ for $x=0$. This model fits quite
well with the data, and explains the stochastically constant behaviour in figures \ref{dN3} and \ref{dN6}.

There is a hidden assumption in this, which is that we have infinite memory to remember
$\phi_n$, and an infinite supply of new $\phi_n$. These finite limits probably explain
when the vertical uncertainty increases for increasing time $s$ in the figures.
In other words, there is a finite set of elements and a molecular
chemistry to word fragments, at least on the scale of the sentence.
This is interesting as it underlines the fact that language is not a
Markov process (with memoryless transition rules). Any cognitive
stream must rely on a finite memory to parse patterns in chunks. This
ought to be obvious on moderate reflection, but it does go against the
traditions of mathematics and physics, which probably explains the
attempts to make Markov models of the artificial order on very large
scales. On the other hand, Markov models can explain the grammatical
classes of the Chomsky hierarchy in a coarse way \cite{chomsky}, so
there is also evidence of the underlying semantic-free dynamics at
work.

When we look at the complement of novelty---i.e. the repetition
frequencies of words, the network effect becomes more apparent.
Singleton fragments are by far the most popular, so most sentences are
basically unique---showing little invariant `significance' within a
document; then fewer and fewer phrases are repeated more regularly
throughout a narrative---apparently forming bindings to a small clusters
of partners to yield repeated `molecular' events. 

So a kind of invariant chemistry with stable meaning occurs at short
lengths. As the lengths of these word-molecules grows into longer
chains, they might acquire greater narrative significance, but they
tend to stand alone and occur only once if one ignores the collective
context around them.  There is a natural cut off at the average size
of a sentence, so we expect the number to be negligible at some number
less than this (around $k=10$), which is shown in the figures.  These
observations suggest that meaning is a scale dependent issue, so it
will involve multiscale complexity. That, in turn, should make us
suspicious of purely probabilistic arguments.  However, at this stage, we are
looking for the `ground state' dynamics of the representational
system, realizing that there will be more to understand later.

In summary, the basic empirical measures that characterize text
confirm a simple sense to the data.  Language is a multiscale
phenomenon, characterized by network bindings on a sentence level, and
phrases clustered at a larger quasi-paragraph level, within documents
of limited subject scope. This points to a basic set of scales to work
with.

\subsection{Random documents}

\begin{figure}[ht]
\begin{center}
\includegraphics[width=7.5cm]{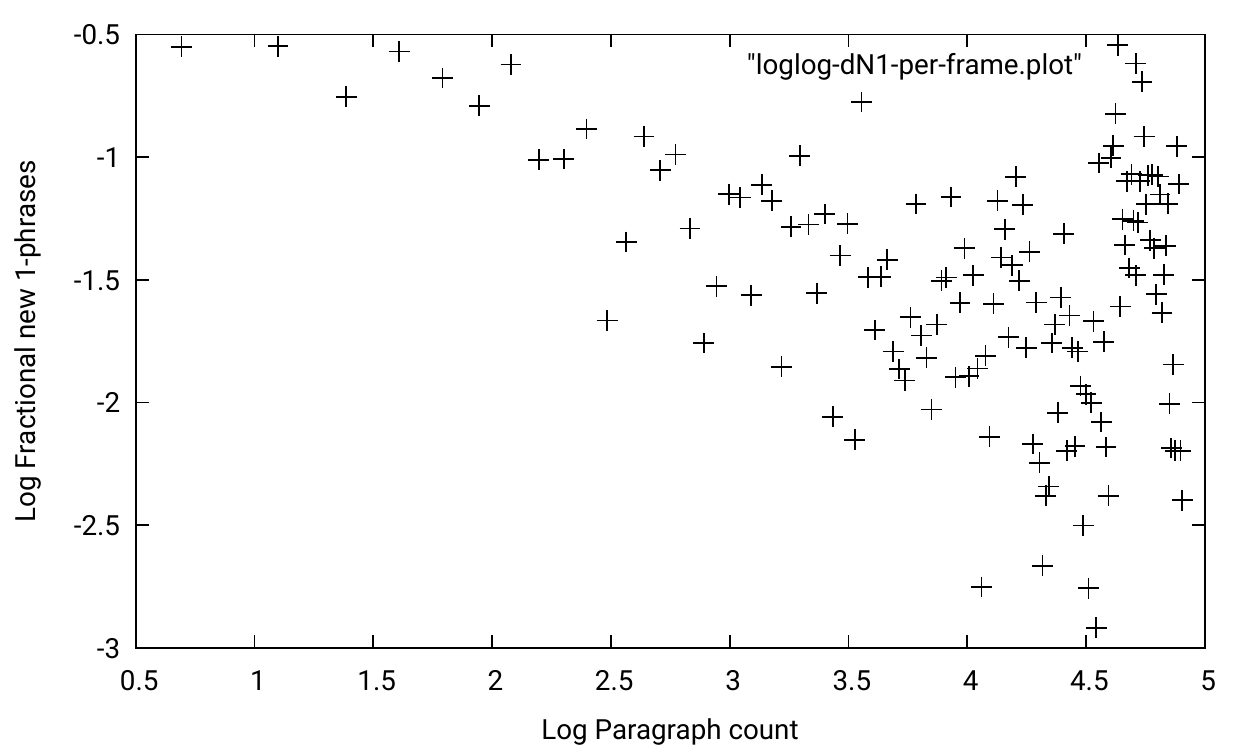}
\caption{\small Comparing figure \ref{loglogdN1} to random sentences:
  we see the hint of the same pattern, slightly though more noisy.
  Since the power form depends on the scaling consistency of the
  process, rather than on an intrinsic property of language itself,
  this should perhaps not be a surprise. However, many authors have
  made the identification of a power law to be a significant
  find.\label{loglogdN1random}}
\end{center}
\end{figure}

It would be remiss not to compare intentional samples to random data. That gives
us a final check on the invariant meaning of meaning!  For this, a few documents were
constructed from sentences sampled from other random documents and
authors. Biases resulting from specific source should then lose focus to noise
quickly. For example, a random comparator to figure \ref{loglogdN1} is
shown in figure \ref{loglogdN1random}.  What's interesting about this
is that the power law pattern is still evident, albeit with a high
level of noise, probably due to the short length. That tells us that
the power law structure has nothing to do with the semantics of
language---at best it's a property of the scaling of syntax after being rendered
homogeneous by local renormalization.

Only the most generally invariant properties of language are likely to
hold up in a summary of a random document. It will be interesting to
gauge whether a summary, extracted on the basis of maximum meaning,
will appear like a non-random document, or like a proper narrative, to
a human arbiter.  Will sentences still feel related as part of a narrative,
when stripped of a cloud of padding?

Other comparisons to random data have used non-randomized subject
matter---only randomized words. Montemurro and Zanette subtract
randomized versions explicitly from the same corpus in
\cite{montemurro1}. Their process of forgetting was therefore on the
scale of words bounded by the scale of the documents rather than
segments of authorship. There are many ways to approach the problem,
but the main thing is that the results are not in contradiction---and that
we should use the conclusions conservatively.

\subsection{Note about decomposition by meaningless bindings}\label{commonword}

For completeness, it's interested to make a cul-de-sac diversion to
explore an orthogonal idea, which relates to the spacetime semantics
of punctuation.  The alternative to splitting sentences by
`sequencing' over random sliding windows, as is used in Artificial
Immune Systems\cite{soma1}, is that we could make use of the promised
semantics that follow the most statistically significant patterns,
namely the privileged meaning of sentences and sub-sentences broken by
the small set of short meaningless words, such as extracted from the
Test1 document\footnote{The occurrence of `city' and `scaling'
  detected here are clearly artifacts of the particular narrative
  chosen, but this should not bias us into thinking that we should
  treat the words any differently. Within that scope, it's possible
  that these are irrelevant words. On the other hand, it could be that
  a multi-scale analysis would reinstate them as being important.}  :
\begin{quote}
\small
it, an, on, or, by, \&, scaling, with,
as, city, this, be, are, for, that, is,
in, and, to, a, of, the
\end{quote}
Breaking on special words is more like the editing of DNA sequences by
identification of gene markers and telomeres\cite{durbin1,baldi1}, so
it has a special interest of its own. However, it assumes that a
mechanism for semantic selection has itself already been preselected over
evolutionary learning timescales, which is like using what we know of
language semantics already. The question is really about how a process
like that could emerge.

The phrases above are derived from the small data samples here.
Results for a larger general English language corpus, on a very large scale, can be found from
the top 10 percent of frequencies from the Google Books library\cite{gwords}:

\begin{quote}
\small
the, of, and, to, in, a, is, that, for, 
it, as, was, with, be, by, on
\end{quote}
In addition, punctuation marks were treated as some of the winners.
Although the list above contains clear winners in terms of homogeneous
frequency within the documents used, there is no obvious chasm in the
frequencies between these words and many other longer words that feel
subjectively more meaningful, e.g.  that are clearly subject-matter
related.

\begin{figure}[ht]
\begin{center}
\includegraphics[width=7cm]{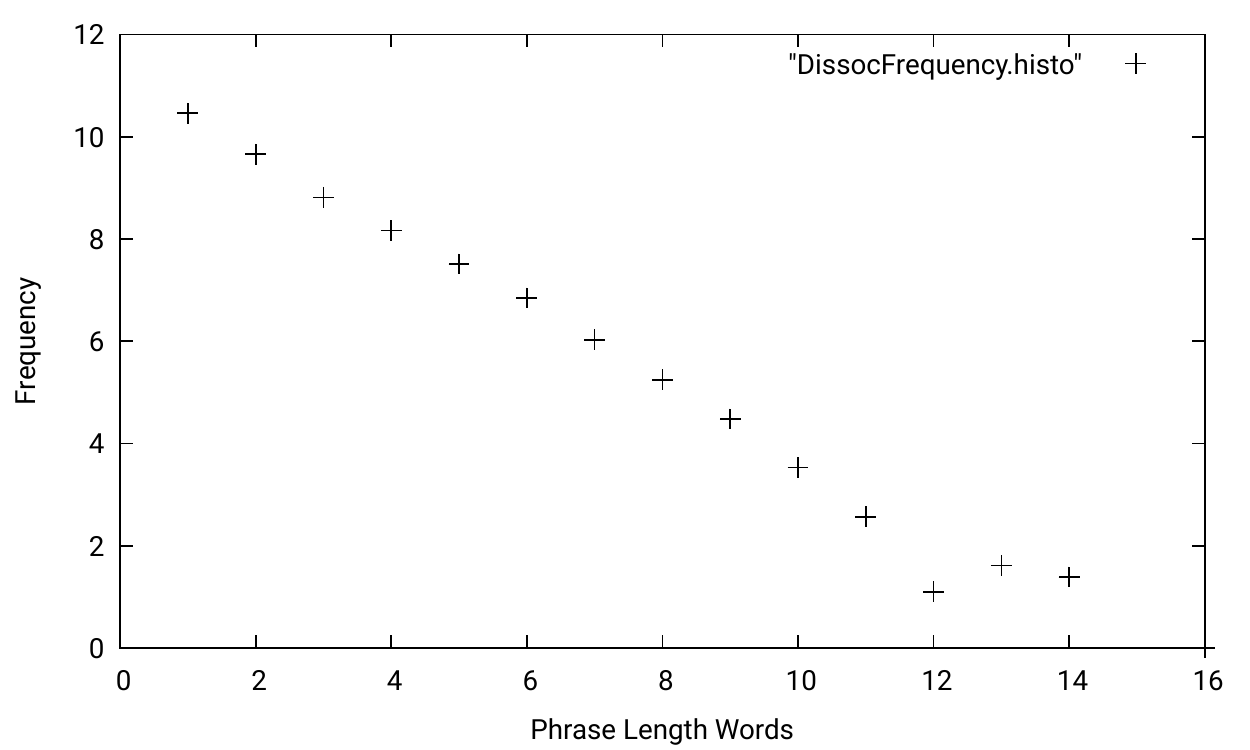}
\caption{\small When sentences are dissociated along meaningless words, the frequency
by length is not less predictable. The fragment lengths are dominated by some short outliers,
and elsewhere there is a roughly exponential decay of fragments by length.\label{dissocfreq}}
\end{center}
\end{figure}

\begin{figure}[ht]
\begin{center}
\includegraphics[width=7cm]{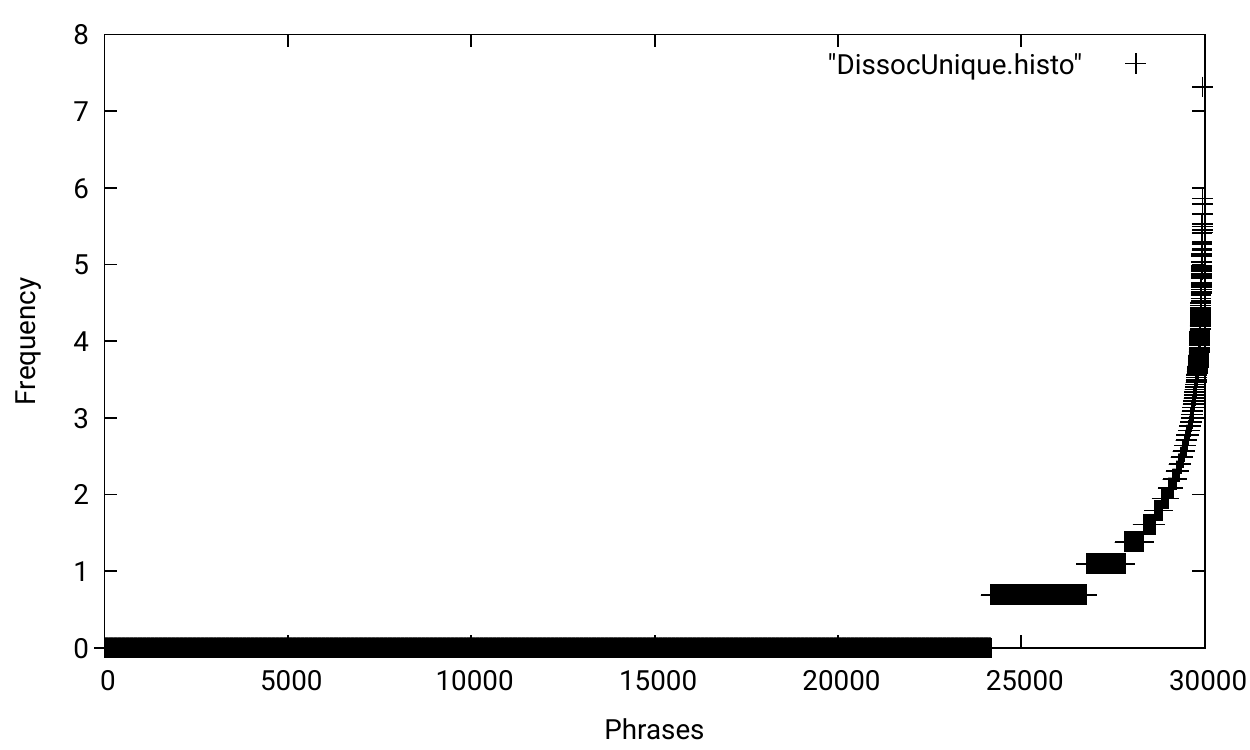}
\caption{\small The uniqueness of `semantic types' shows that the specificity of sentence fragments
makes their statistics highly irregular. A few trivial phrases dominate, but most phrases
only get used once, or perhaps a handful of times.\label{dissocunique}}
\end{center}
\end{figure}

From a learning perspective, what effectively makes the words
irrelevant is rather the effect of cancellation by `signal
interference'---scanning a stream in this way is effectively about
building a kind of symbolic interferometer that sharpens our notion of
meaningful change\cite{interferometry}.  Only more extensive
interference of signals over long times could possibly select a
smaller set\cite{gwords}, to eliminate the issue of
alignments\cite{durbin1}.  In the case of punctuation marks, there is
a contextual identification of status as a marker, as in chemical
markers in DNA. These may be based on certain promises (or lack
thereof), but ultimately they are selected by a process that
discriminates context. See figures \ref{dissocfreq} and \ref{dissocunique}.

This process of elimination could be called the Sherlock Holmes approach to meaning.
When broken by these meaningless words, what remains appears superficially
meaningful (as Sherlock Holmes deduced---what remains after eliminating
the banal must be the significant in some sense).
On the negative side, a strategy of splitting along these words may eliminate potentially
interesting concept names, where they play a role in binding of concept names:
\begin{quote}
\small
murder by knife\\
type of car\\
to die for\\
to live by
\end{quote}
Although statistics alone would not see this from a small sample, it is a function
of the fact that some phrases are not specific to a single narrative---even specialized language
is embedding within generalized larger language, which is ignored at our peril.

Clearly, statistics alone (even very good statistics) are insufficient
to discriminate the proper role semantics of words. What's missing is
a form of contextual selection as we know from evolutionary
processes\footnote{In any promise binding there are both promises to
  give (occurrences) and to take (selections), called + or - promises
  in Promise Theory parlance.}. This brings us back to the negative
selection processes that focus effort in immunology. The next stage of
recognition would be to account for the promises of the words, i.e.
their functional roles learned by language-evolutionary timescale
selection.  There is some inevitable arbitrariness here too in
selecting the set of words along which to make breaks. It indicates,
again, that quantitative methods alone cannot explain the significance
of text---context is required, in the form of clusters of named
concepts, not just amounts.  Learning on multiple timescales is
involved in establishing the contextual parts of language signalling.

\section{Spacetime measurements}

\subsection{Coordinatizing for stream measurement}

In dynamical terms, a coordinate system is simply the outcome of a
cyclic calibration process, which other processes are compared to
in order to `race them' against one another and give relative
comparisons the flavour of local absoluteness. Coordinates enable us
to distinguish and label information occurrences reliably against a
common standard. No measurements can be given meaning without some
form of coordinate system.  The dissociation of narrative into
repeated fragment processes is what provides this calibration for
document parsing.

A next step beyond simple interferometry to coax alignment is to try to 
rank process anomalies in the `spacetime measures' (analogous to distance and time)
for importance (see section \ref{meaning}). 
Qualitatively, we experience events as meaningful if they occur within
a limited scope (typically more than once), not merely scattered like
dust or background noise.  To measure how homogeneously distributed
words are, we need another memory process that works to measure local
occurrence rates---total occurrences are not sufficiently precise.  A
uniformly distributed measure (on a metric space) is defined as one
for which the number of occurrences within an open ball depends only
on its radius and not on its centre. In other words, if we draw a
ring around a region in one location, we find approximately the same
number of samples as if we move the ring to another location.

There are other measures of inhomogeneity.  Standard statistical
measures, like symbol entropy for instance, turn out to be largely
inappropriate for this local identification because they computed over
entire corpora and all their process paths, and are too dependent of many assumptions about
scale and classification of types over complete systems.  The problem
with such definitions as entropy is that, as seen earlier, longer
meaningful phrases are rarely repeated more than once is a single
source. As we add specificity and carefully described distinguishable
things to a story, new information inevitably is what characterizes
the stream, so we can't use a statistical basis for identifying within
a story. We've already strayed into the domain of general language, in
which stories borrow from one another over time.

If one tried simply to count fragments of text as semantic `types',
i.e. by role, over an entire document, then we would not be able to
distinguish their locality, which is crucial to their overall meaning.
Context is a local network effect.  We have to break the document
`spacetime' into parts that can be compared and contrasted.  This is
because stories are partially order dependent---they aren't just
random sets of symbols like a Scrabble set where we might count the
number of similar letters hoping to find a difference between one set
and another.  So we have to be able to distinguish different
occurrences of each signal utterances from one another on the scale of
compositional molecules too.

Fragments within a single sentence are not strongly ordered (one can
change the order and still read the sentence in many languages, with
some caveats). The order of sentences and paragraphs is possibly more
meaningful, however. One expects to see trend patterns on a larger
scale more clearly than by focusing on words alone.

The variability of events as measure by length, and counted in words,
indicates that we should consider text streams as an inhomogeneous `non-equilibrium'
problem, which therefore requires a variable sampling regimen in order to capture
the non-linearity from effective boundary conditions interpreted in terms
of Nyquist frequency\cite{shannon1,cover1}.

\subsection{The metric basis for significance (anomalies)}\label{meaning}

Finding a coordinate system by which to measure a story is a matter of convention in a
given language; but, for the Western Germanic and Romance languages,
at least, we have the convention of paragraph, sentence, and word as
the compositional scales. In order to distinguish localization of
concepts, there would then be an imperative to keep each paragraph
short, even at a standard length---but that shouldn't be assumed from
the outset.

The suggestion that emerges from these simple studies is that the
subject of a stream of text is likely to be developed within a
framework of patterns which are repeatedly identifiable in localized
clusters, and those clusters form progressions throughout the stream
until the end. We needn't know anything about language to find those
patterns as long as we can distinguish them, so it's to be emphasized that {\em no prior knowledge of
  language or meaning is involved} and this method could be applied to
any other kind of data stream in principle.  This suggests that any
time-series might be resolved by a set of pseudo-periodic functions,
which can be used to span the normal background and reveal anomalous
occurrences. Indeed, this was the approach used for other quasi-periodic processes
in \cite{burgessDSOM2002,burgessC14}.

The paradox of meaning is that it's a fiercely scale-dependent issue.
If you try to look at what stands out over a small amount of data, on
too small a scale, every new string appears significant.  In a
narrative, by contrast with a steady state immune process, the
development of new fragments is not a short memory Markov process but
almost every new fragment is new at the scale of the story.  On the
other hand, repeated concepts, over a large scale, must always become
insignificant in some asymptotic limit. Statistics is good at finding
signal in noise only when there are simple invariant semantics in
play. Once there is an explosion of semantics, as in storytelling,
statistics probably have a basic handicap in uncovering any trace of
the story. As in particle physics experiments, one has to know and
understand the story quite deeply beforehand in order to test it with
statistical measurements.  What this suggests is that language without
fixing scale is meaningless.  Language must break the scale symmetry
in order to represent context, and thus pure probabilistic
representations are vacuous of intentional content. We should still use
dimensionless representations to match language representation to environment
representation however---that's a different issue. Each representation
will nonetheless have its own `proper time' scale.

Could we, for instance, identify the most important phrases within
each paragraph and choose these to be representative of the purpose of
the narrative in that paragraph, creating a semantically scaled
hierarchy of meaning?  This is not quite like the frequency analyses
one performs in code breaking.  It would form the basis for
signposting along the storytelling journey, and the remainder of the
text in each paragraph could be considered explanatory attribution
that bolster that central message.  It's such a simple hypothesis that
it seems to be worth testing: does each paragraph promise a `single'
nugget of meaning?  Similarly, could we detect legs of a document by
looking for ``change of topic'' or change of emphasis?

Arbitrariness in our choices choice might be both compelling and
unsatisfactory as we search for the overarching story about what makes
stories into good stories. However, we can try to postulate a model
based entirely on localized spacetime {\em distinguishability}. 
For phrase $\phi_n$, where $N_D(\phi_n)$ is the frequency within a
document $D$, then we can postulate a relative meaning $\phi_i$ within
the document of the form:

\beq
\mu(\phi_i) \propto \frac{|\phi_i|}{|\Phi|} \times f\left( 1 / N_D(\phi_i) \right)
\eeq
for some function $f(\cdot)$ of the relative frequency, for length in
words $|\phi_i|$ and some fixed scale on the level of sentences or
paragraphs $|\Phi|$.  A problem with this definition is that it
doesn't take order and distribution into account, as no purely
frequentistic study can, It can't distinguish twenty references to a
word in a single paragraph (making it obviously interesting) from
twenty random references throughout a document.

However, we could do better: there are two aspects to measuring significance:
\begin{itemize}
\item {\em Transients}: For a uniformly distributed measure (i.e. one in which the count inside a certain region
depends on the size of the region, but not the actual time itself)
\beq
\frac{dN(\phi)}{d\tau} = -\lambda = \text{const}
\eeq
where $\tau$ is the proper time for the document. Transients in $N(\tau,\phi)$ reveal meaningful words.
This suggests that {\em dynamical properties} of the language reading process
play a role in importance.

\item {\em Effort}: There is also a `work or energy' argument to meaning. The amount of work
an agent puts into expressing an idea could be considered proportional
to its importance. We know, from regular and irregular grammatical
phenomena, that the cost of irregularity is generally borne only for
ideas of special significance with enough importance to sustain them. Other words
are quickly regularized and rendered as generic patterns. A first-step hypothesis
to capture this on a linear stream would be  to make the length of a fragment
$\phi$, in words, signify its importance. This might be generalized later to
other scale dimensions in a multi-scale refinement.
\end{itemize}

The suggestion then is that a basic importance would have the form:
\beq
\mu(\phi_i) \propto \frac{{|\phi_i|}}{|\Phi|} \times \left( \frac{d N(\phi_i)}{d\tau} \cdot \xi \right)^{-1}
\eeq
where $\xi^{-1} \sim $ is the average rate of sentence or phrase
production over a typical memory decay time.  One could eventually go
further into specifics: it's unclear whether this purely flat learning
measure would be sufficiently sensitive to clusterings. One could also
add some non-linear feedback amplification to the memory circuit by
adding a cache of recent terms, which is decays more slowly that the
word memory---to follow the trends on a paragraph scale. This would
provide a persistent context at least within a single narrative.

\beq
\mu(\phi_i) \propto \frac{{|\phi_i|}}{|\Phi|} \left( \frac{N(\phi)}{N_0}\right)^\beta \left( \frac{d N(\phi_i)}{d\tau} \cdot \xi \right)^{-1}\label{eq1}
\eeq
for some constant $N_0$ of the narrative or document, and power $\beta$.
With this addition, one can imagine an iterative multi-pass reading of every document to familiarize
the system with the meaningful terminology---like skimming a document before reading
it to get a gist and know what to look out for. This is a natural strategy for any
learning system: a simple adaptation to turn non-supervised learning into a self-supervised
learning, by trying to geometrically stimulate meaningful selections. Each new reading
reflects the document back in a mirror of one's previous assessment.

The satisfactory aspect of this measure is that it results in a natural
way from the competition of rates between real-world scales and finite internal resource limits.
There is, however, the question of how stable this importance ranking is. If it were too stable,
one would have eliminated the importance of context. If not stable enough, everything would be
recognized as anomalous and important. This is the challenge of selecting an appropriate
equilibrium between signal and noise.

An effect which is normally dismissed in machine studies is the effect
of `emotion' on perception and cognition, although its importance has
been appreciated for a long time\cite{emotion1,emotion2}.  Emotion is
a blunt classifier of context, which provides a quickly computable
bias for characterizing situations in which information arrives. In
short, it works as a quick index classifier for experiences.  To
remove the stigma of human weakness, Kahneman calls this fast and
blunt characterization `system 1', whereas he calls rational reasoning
by semantic storytelling `system 2' \cite{kahneman}.  This kind of
quasi-emotional response is a candidate for a possible co-activation
signal, as it amounts to an independent interpretation of information.
In this sense, the information theoretic viewpoint would claim that all the
information was already in the source, so multiple readings would yield anything
new. However, in a promise theoretic sense, it all depends on how the information
is perceived.

\begin{figure}[ht]
\begin{center}
\includegraphics[width=6cm]{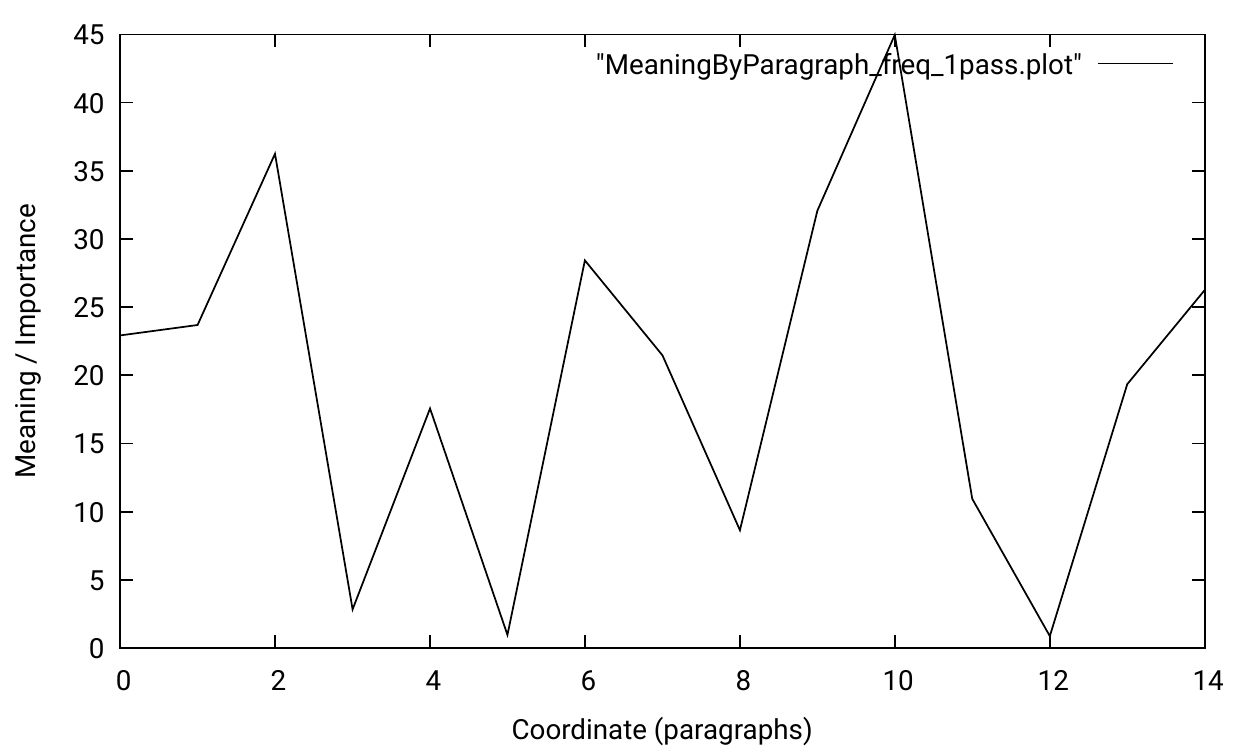}
\includegraphics[width=6cm]{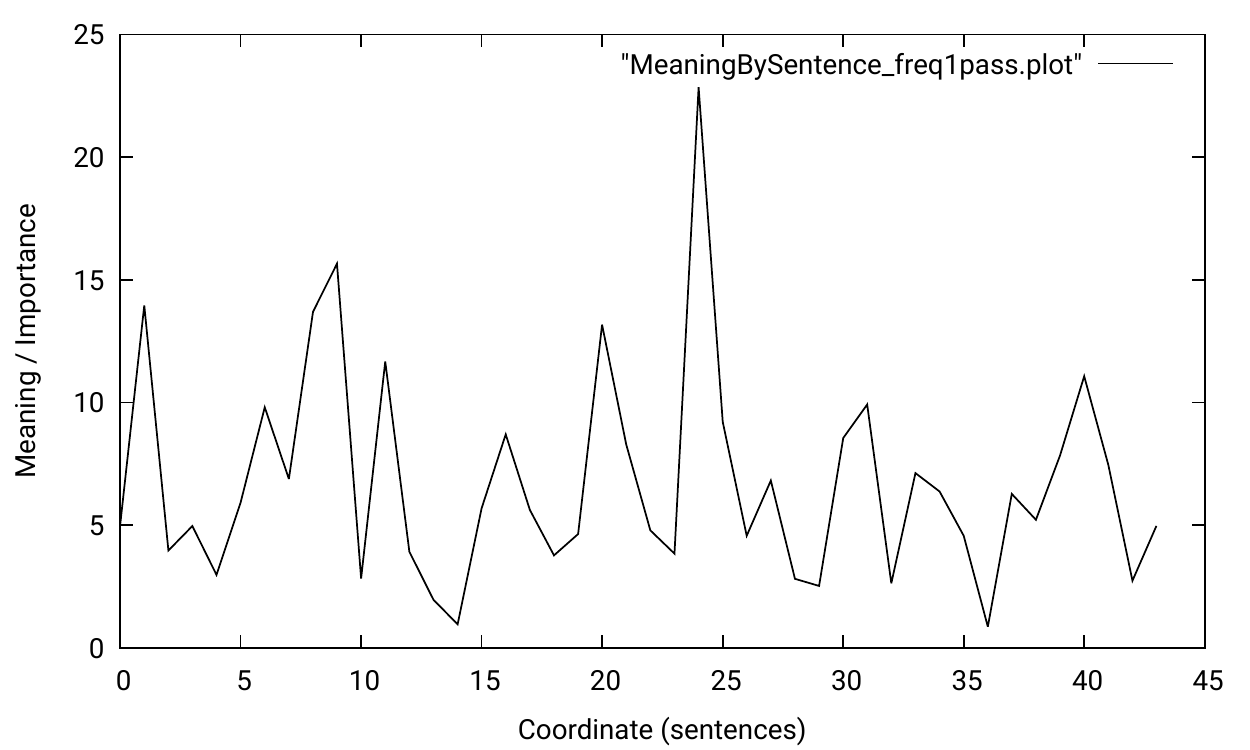}
\caption{\small Single pass cumulative learning and assessment of
  meaning from a small document, assessing coarse-grained paragraphs
  (above) and sentences (below). The results show that some paragraphs
  are assessed as being more important than others in a way which is
  not biased by counting of location from start to finish.  Owing to
  the sparsity of points, lines are used between data points to
  make the graph easier to read, leading to the witch's hat effect
  which is exaggerated for paragraphs as there are fewer.\label{MBP1}}
\end{center}
\end{figure}

\begin{figure}[ht]
\begin{center}
\includegraphics[width=6cm]{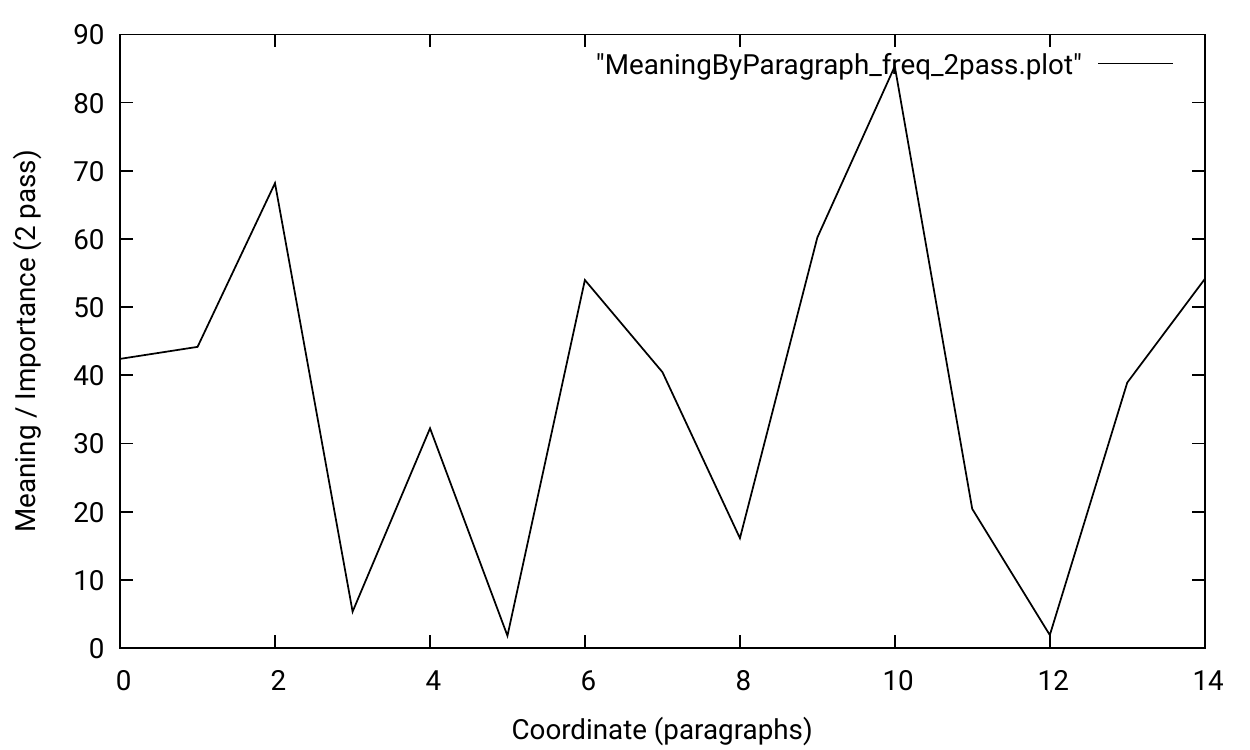}
\includegraphics[width=6cm]{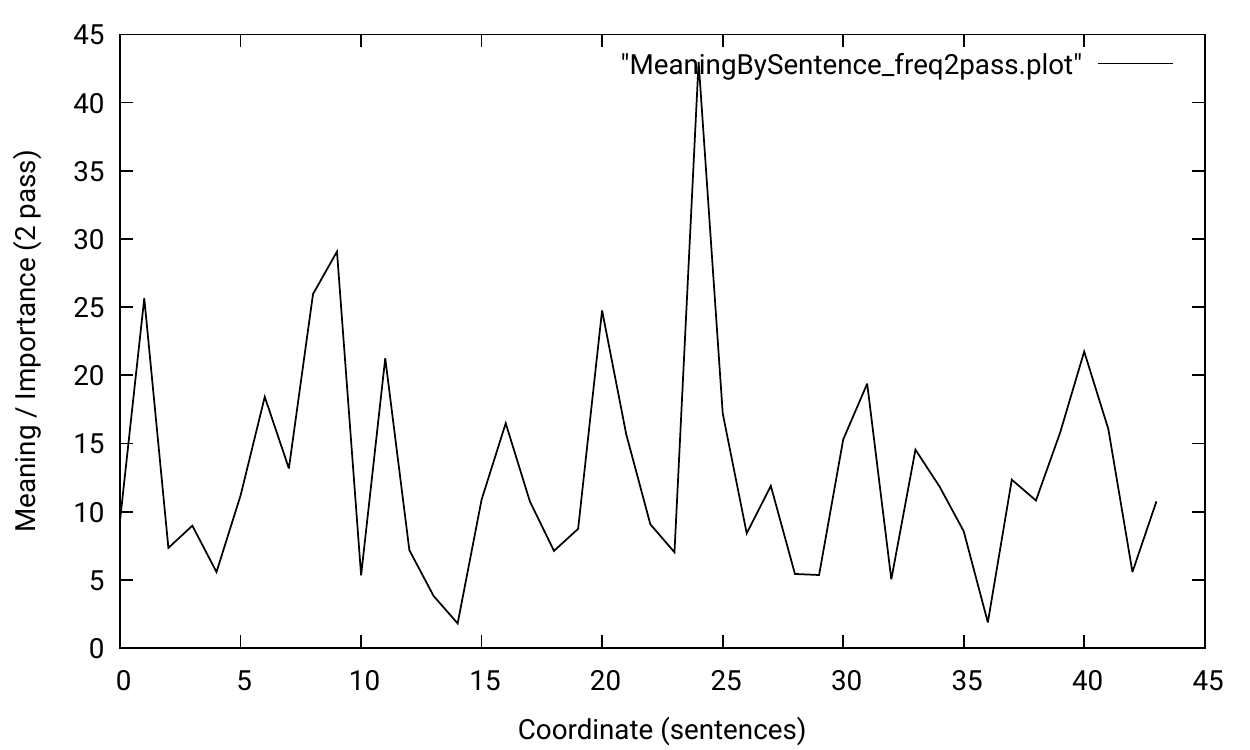}
\caption{\small (Compare to figure \ref{MBP1}) Two pass cumulative learning and assessment of meaning from a small document, assessing coarse-grained paragraphs (above) and sentences (below).  This shows that a single pass is qualitatively similar to a double pass, so adding more learning does not make a large difference to the unbiased process, because it is already tuned to forget faster than most documents.\label{MBP2}}
\end{center}
\end{figure}

\begin{figure}[ht]
\begin{center}
\includegraphics[width=7cm]{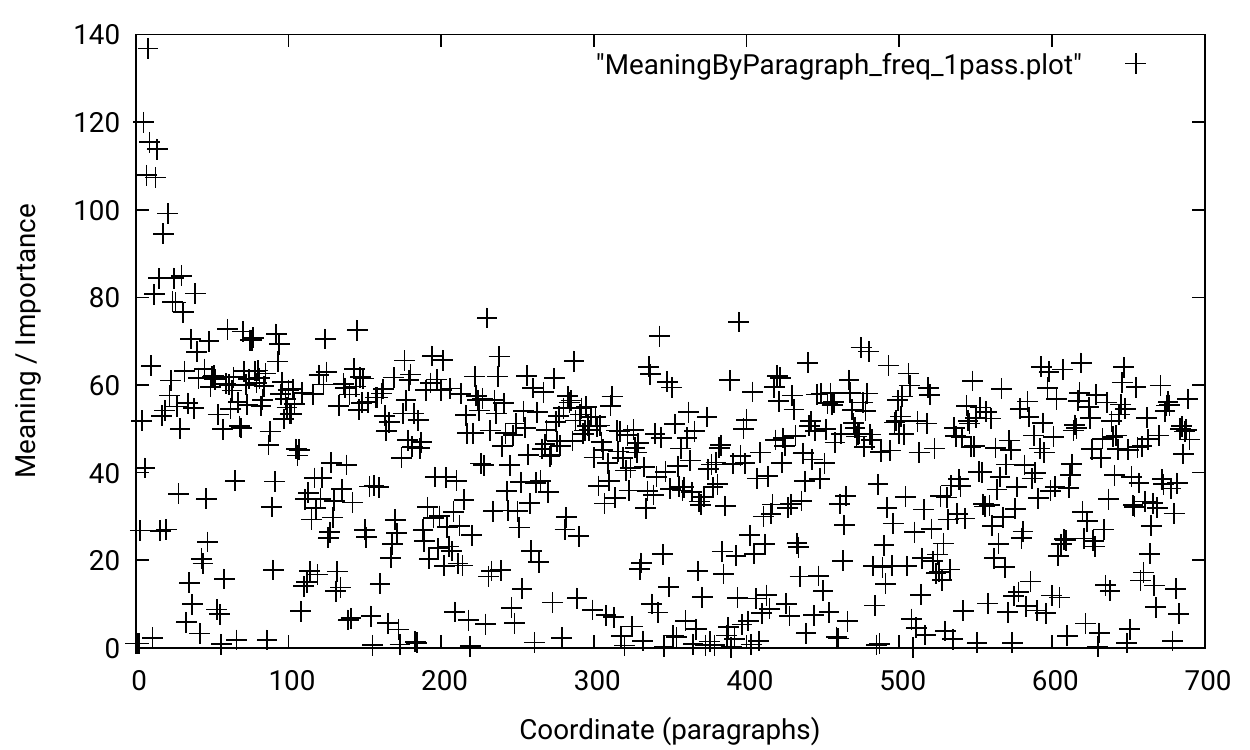}
\includegraphics[width=7cm]{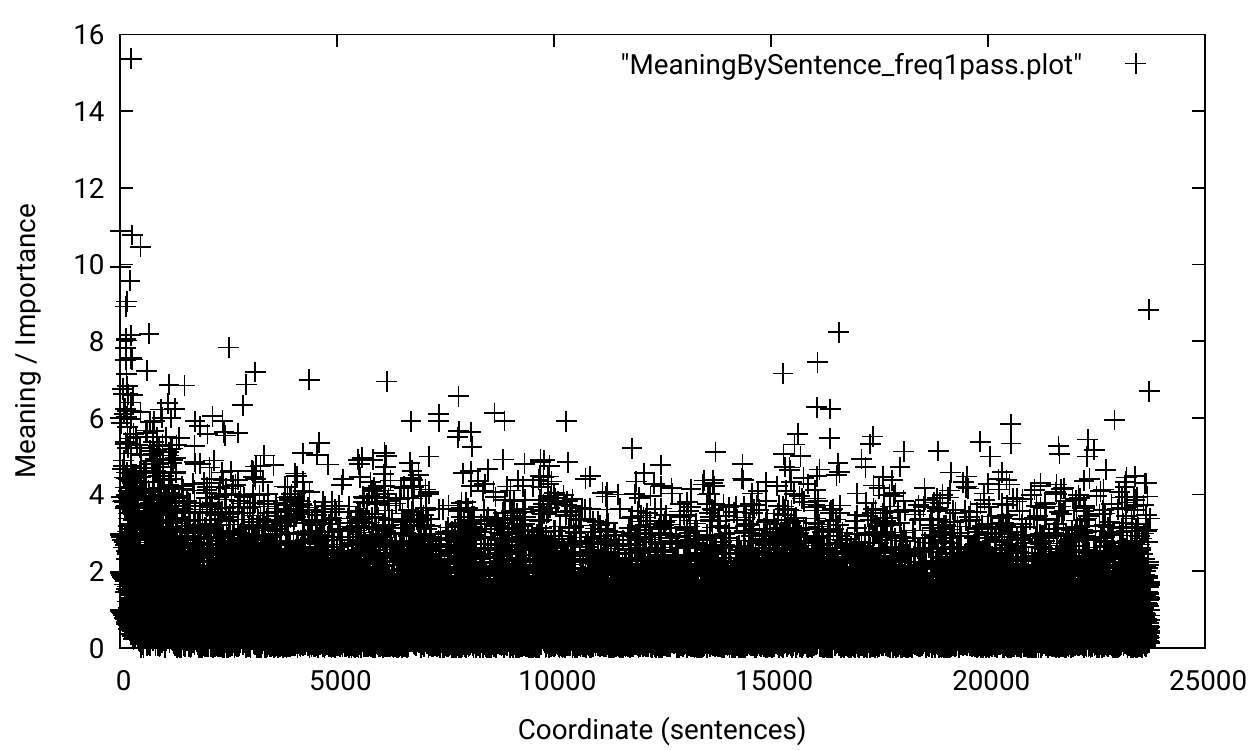}
\caption{\small Single pass cumulative learning and assessment of meaning from a larger document (Slogans), assessing
coarse-grained paragraphs (above) and sentences (below). A much higher density of points makes
the variability of meaning assessment difficult with the naked eye, so success of failure has to
be judged by human arbiter.\label{MBP1slogan}}
\end{center}
\end{figure}

\begin{figure}[t]
\begin{center}
\includegraphics[width=7cm]{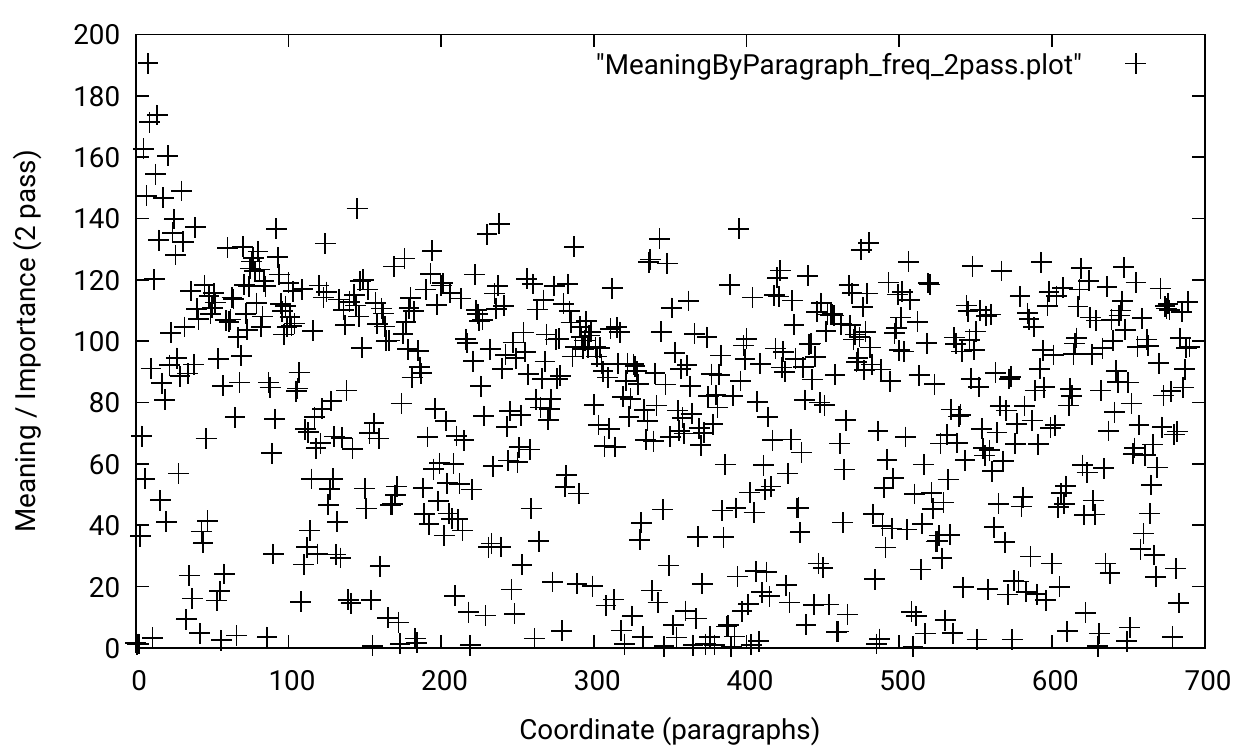}
\includegraphics[width=7cm]{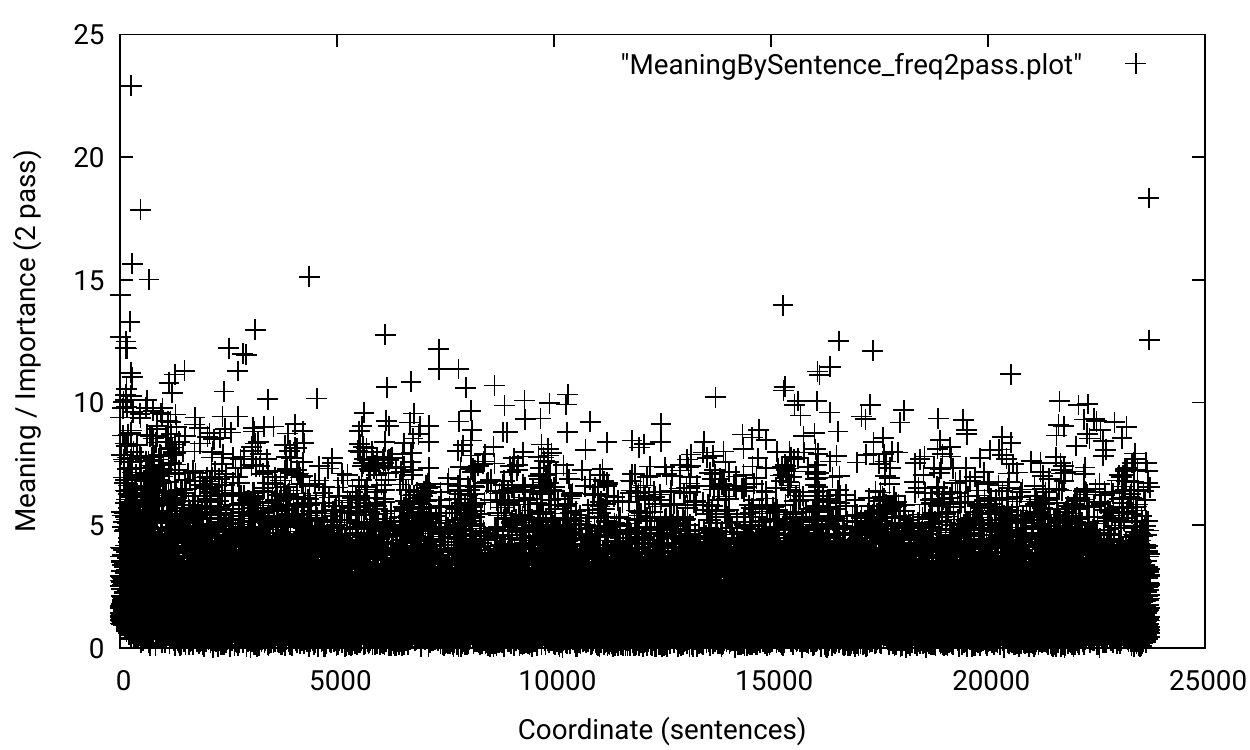}
\caption{\small Two pass cumulative learning and assessment of meaning from a small document, assessing
coarse-grained paragraphs (above) and sentences (below). There is no significant
quantitative difference between this and figure \ref{MBP1slogan}, and the qualitative result
was equally vague, suggesting that the tuning of scales made repetition inconsequential.\label{MBP2slogan}}
\end{center}
\end{figure}

Figures \ref{MBP1}-\ref{MBP2slogan} show the behaviour of an
approximation to equation \ref{eq1} make with pseudo-periodic learning
functions, for short and long documents. One could rank regions either by
paragraph, leg, or by sentence (or the product of the two to capture two
scales). In measurements, the attempt to suggest a logical `AND' of 
dependencies for a dependent outcome did not improve significantly
over the result for sentences alone. Other factors seem to make the
slow variation less important on a quantitative level. Later, a
different approach was used to select away sentences entirely
rather than rely on quantitative measures---indicating that a
promise duality of offer-acceptance is the most effective way
of selecting importance. That's suggestive that an immunological
approach to selection is more helpful than a spacetime
approach based on measures of average distance and time (in words).
However, other results also point in the other direction.

Another question is whether the learning approach to gathering realtime
statistics is sensitive to transients or data ordering issues.
In any small amount of data, one might expect a transient response at the
start, in which the system goes from a blank slate to a certain level
of recognition, but the rate at which this dissipates depends on the
the relative timescales. On a second pass, this transient should have
been washed away by the steady-state solution. Comparing the two
passes, however, shows that the choice of parameters is satisfactory
in eliminating obvious artifacts of this kind.  Figure \ref{MBP2}
(dual pass) can be compared to \ref{MBP1} (single pass).  Apart from a
tiny reduction in the initial transient decay, there is not much
difference between them for small data. In larger data, the story is
similar, so there seems to be little benefit to prior training as far
as this measure alone is concerned.

We can further check whether the results are stable to multiple passes
over the same data (especially to check whether there are any
intrinsic biases due position from start to finish that might occur as
transients or artifacts of the linear forgetting). Again, in practice,
there was no effect on the approach. The reason for this (possibly
surprising result) is that there is an implicit rate of forgetting in
rate-based dynamical method. The machine learning approximation
described below basically eliminates the effect of persistent
transients in the absence of repeated stimuli.  Small differences
could be seen between the single and double pass results in specific
weights, but the overall effect on a multiscale analysis was
insignificant.

In summary, the spacetime approach seems to provide quite stable
results when tuned to the `natural frequencies' of the most
repeated patterns on the smallest scale of the stream $\phi_1$.
It's worth emphasizing here that the ranking of meaning or importance
has, thus far, only been shown to have reasonable properties on the
data studied. It has not been shown to agree with any common sense
interpretation of meaning by human judges. That remains inaccessible
to the current method, or alternatively yet to be shown by heuristic
inspection. The milestone here is to ensure that a consistent measure
of importance can be formed entirely from the scales of the narrative
process, by counting symbols across a range of scales.  The method
indeed behaves properly and scales in an expected fashion.

\subsection{The scale or scales of meaning}

\begin{figure}[ht]
\begin{center}
\includegraphics[width=7cm]{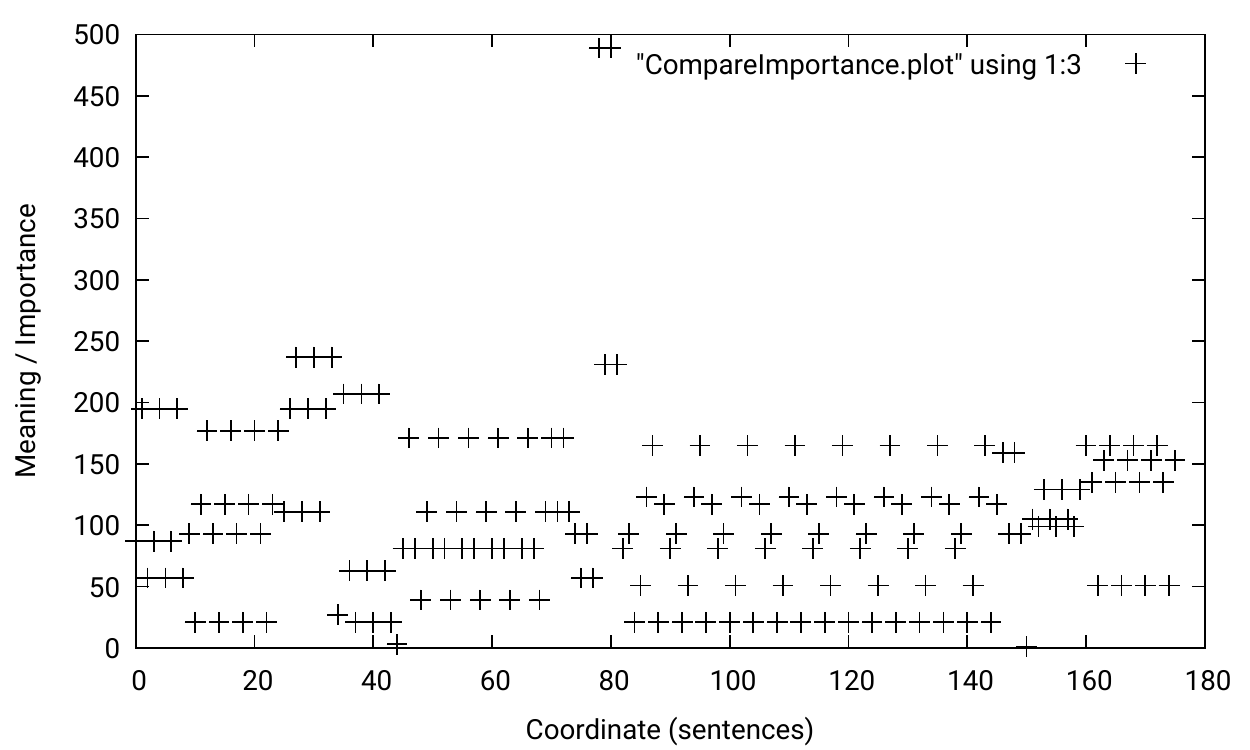}
\includegraphics[width=7cm]{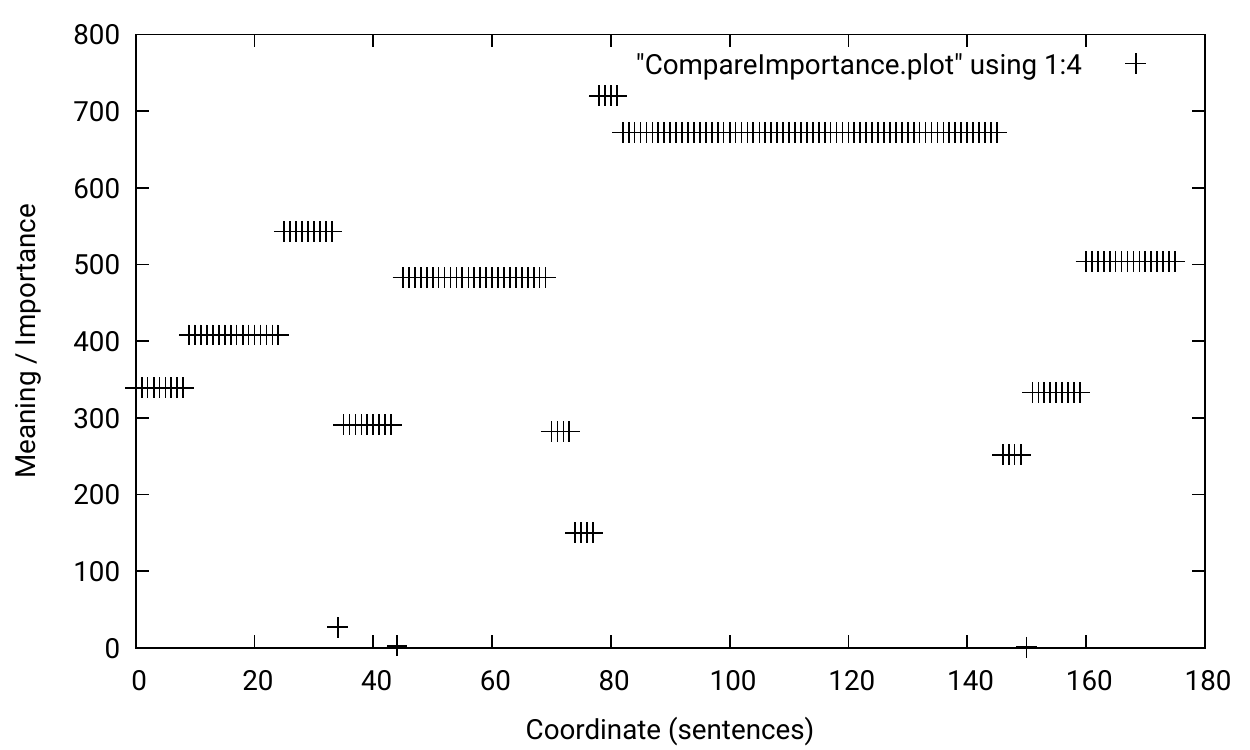}
\includegraphics[width=7cm]{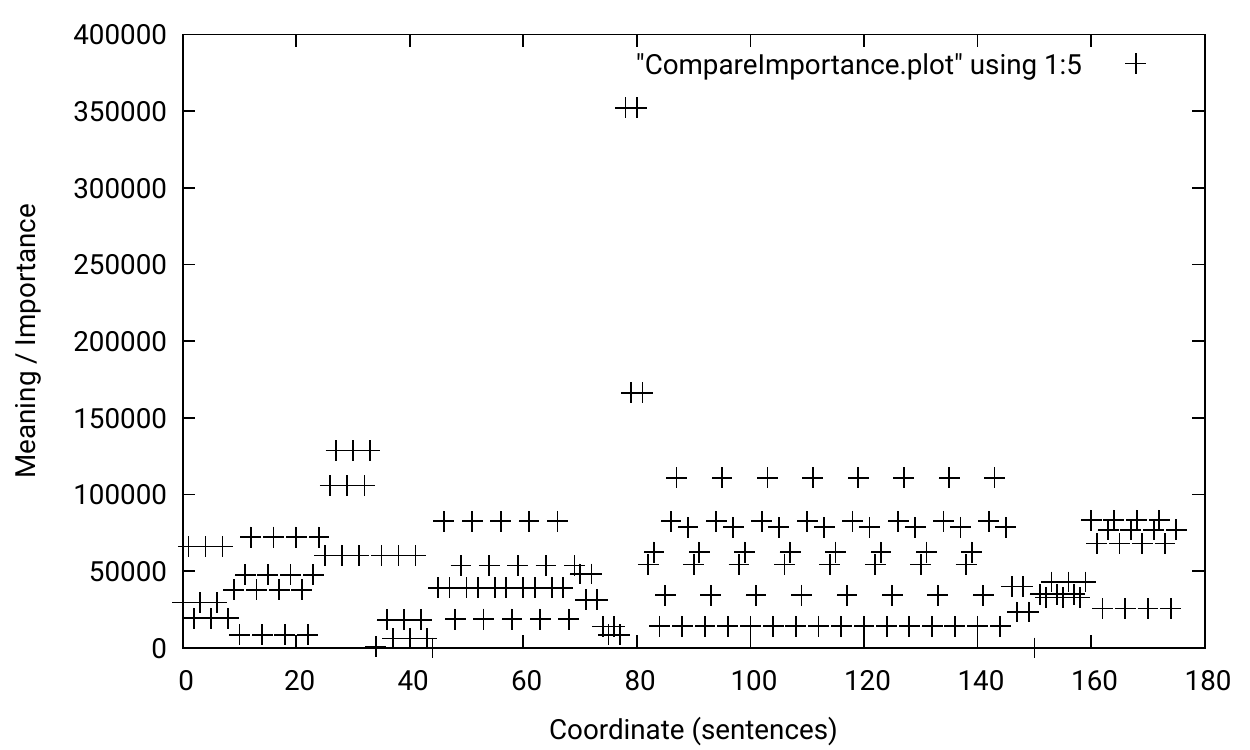}
\caption{\small Combining the significance ranking of a short document
  by examining rank by sentence, paragraph, and the product of
  paragraph and sentence (from top to bottom). Apart from the vertical scale change, subtle
  but notable differences arise in the rankings, but there is no clear
  result to indicate how paragraph ranking plays a role from a purely
  quantitative measure like this, as paragraph distinctions are
  ambiguous.\label{SPrank1}}
\end{center}
\end{figure}

\begin{figure}[ht]
\begin{center}
\includegraphics[width=7cm]{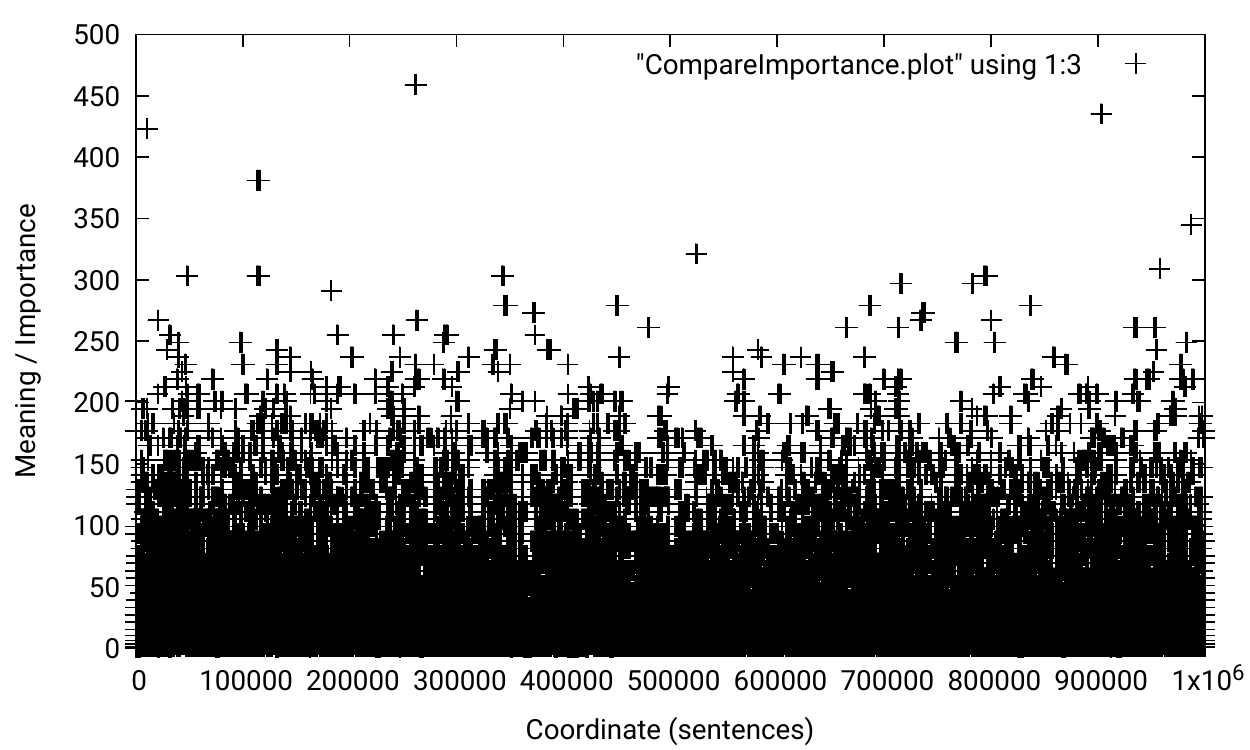}
\includegraphics[width=7cm]{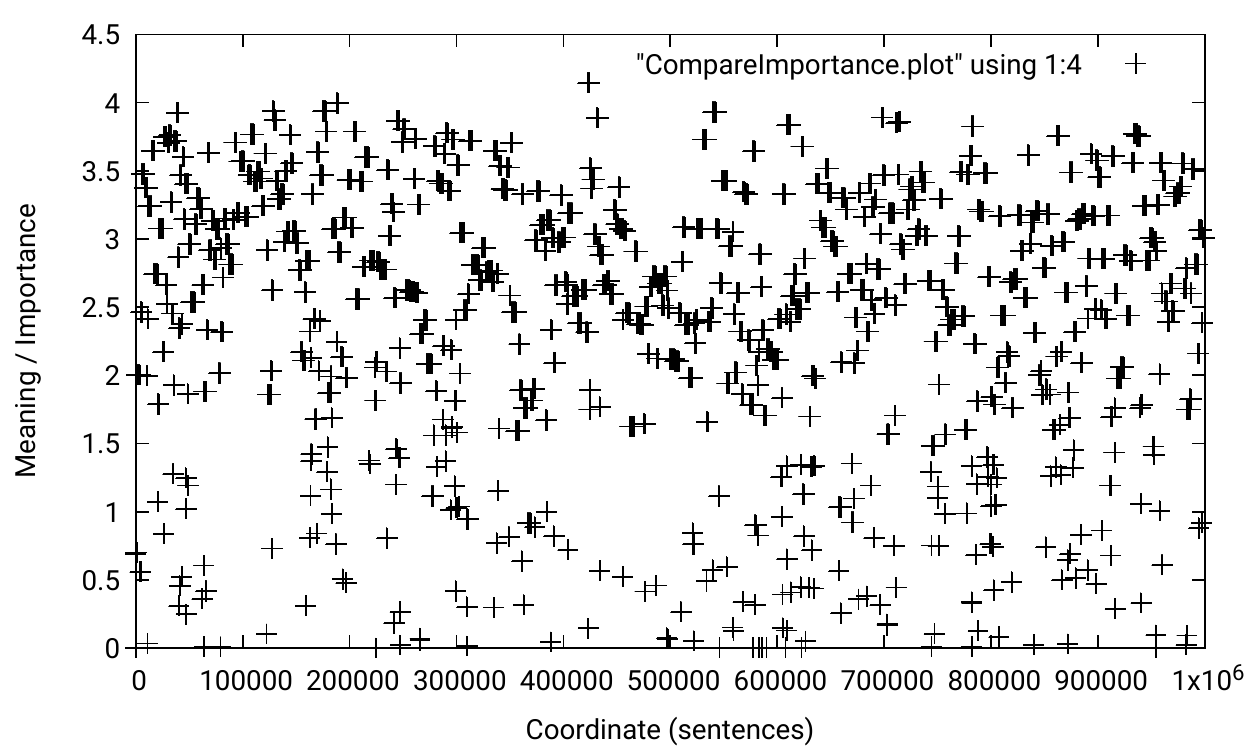}
\includegraphics[width=7cm]{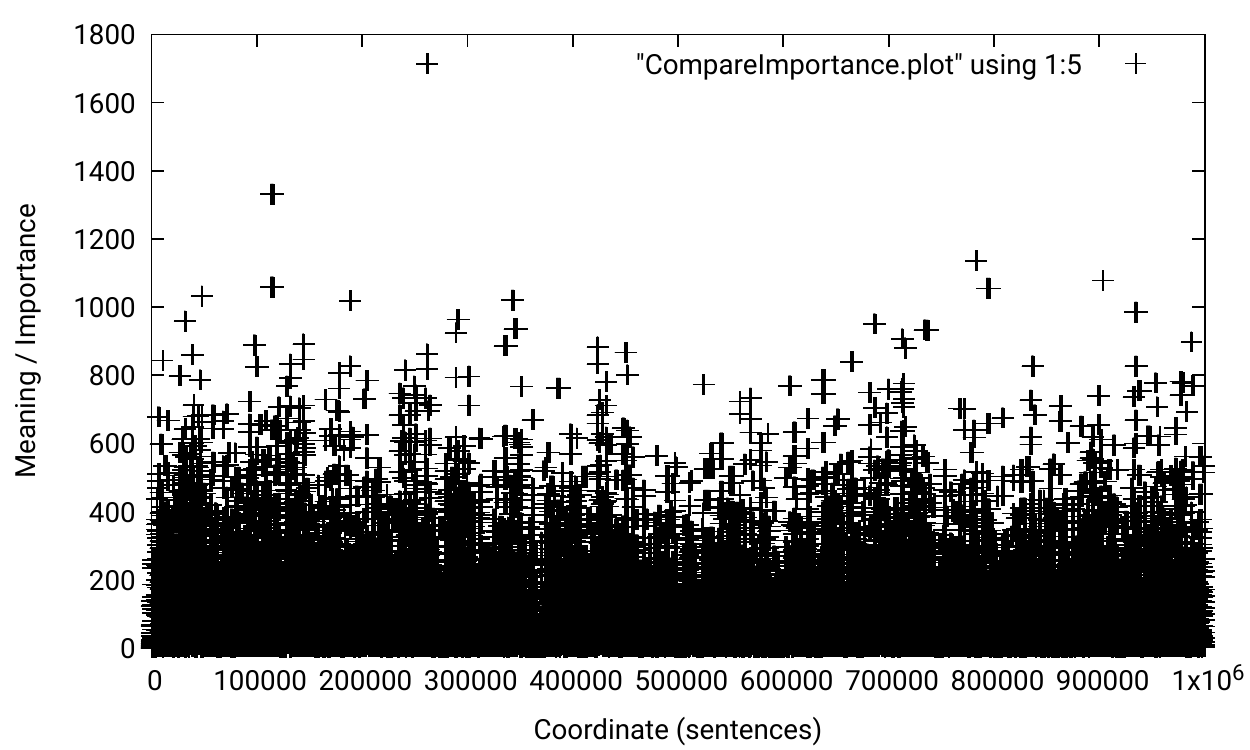}
\caption{\small Repeating the ranking sentence, paragraph, and the
  product of the two from figure \ref{SPrank1} to a longer document
  (Slogans) by sentence. The effect is harder to perceive at scale,
  but nevertheless present. The goal is to choose a timescale over which
the process is smooth and constant over the smallest frequency changes.\label{SPrank2}}
\end{center}
\end{figure}

Since scale plays a clear role in the way interferometry of measures
works, in practice, it's worth taking a moment to look at the effect
of ranking by combined (dependent) measures across two natural scales:
the sentence event, and the longer `leg' or pseudo-paragraph. Does
such an approach make sense?  As a further check on the scaling
properties of the importance ranking, then, we can test the method on
larger documents to see the distribution and discrimination thresholds
that yield stable outcomes for sentences and paragraphs on rather
different kinds of longer document.  On a large document, the graphs
(see figures \ref{SPrank1} and \ref{SPrank2}) become somewhat harder
to understand.  That is to be expected, since we can't predict where
the most meaning is to be found in a document.

If we attempt to define meaning quantitatively, by reducing semantics
to membership in a particular probability class, then this forces us
to define a numerical threshold that marks the line for discriminating
acceptance of non-acceptance of that test\footnote{This is a trick
  used often in scientific literature, to sweep a difficult issue
  under the rug. In the field of semantics, one really can't get away
  with this sleight of hand however. We have to be quite clear about
  what is being promised.}. The necessity of resorting to thresholds
arises from the fact that we've suppressed a clear semantic
distinction into a blunt quantitative one, without identifying a scale
to calibrate the distinction.  The threshold is effectively the echo
of names that were hidden.  Arbitrariness of naming is turned into a
corresponding arbitrariness in numerical value. This issue is both
deep and subtle, and while it's often used in scientific literature to
pander to believers in the primacy of quantitative measurement, it
returns here to haunt quantitative methods in various guises.
Ultimately, all such arbitrary scales must be rendered by a neutral
third party, which can calibrate them according to its single
standard. Any remaining thresholds that can be viewed as ratios point
to critical phenomena, i.e. phase transitions in the model.

The unavoidable conclusion from these measurements is that meaning
cannot ever be considered to be a purely statistical scale-invariant concept, it has
to be contextualized by phenomena that can be used as measures. The
power law scaling of the type of graph that emerges from such a
process is at least consistent with the statistics seen here in figure
\ref{dN1}.

\section{Narrative discrimination based on continuous memory decay rate}

Based entirely on the scaling considerations thus far, it appears
clear that we have to embrace the {\em dynamical aspects} of narrative
process when looking for significance. Static statistical ranking
is not a guide to meaning when communication is viewed as a process. Transient phenomena play an
important role at the scale of words and sentences, and running
assessments of clustering have an average effect to lift up the
attentiveness to capture these clusters.  Long scale variations are
thus a boundary condition on the interpretation of signal on a smaller
scale.

Some readers might feel this is peculiar. From linearized single scale
thinking, this would not make sense. The average information measured
on a coarse grain scale is the same information as the data that went
into it. But the reason that does not apply here is straightforward:
when focusing on specialized narratives, rather than linearized
corpora, there are many dependencies that bring clustering and
non-linear feedback to the counting of phrases. These do not average
out over all process dynamics, as they would for a coarse graining
over intentionality too. Here, we can find more limited meaning where
it exists (just as primitive organisms would) in the spacetime
characteristics of a single coherent process.

\subsection{Machine learning approximation to process scaling}

A simple way to embed this principle of distinguishing anomalous
inhomogeneity from regular homogeneity into learning is to allow
memory of phrases to decay linearly with time, and refresh them each
time they are seen anew---mimicking a pseudo-periodic function for each basis phrase.  This
is effectively very like  `immunological learning', seen in disease immunity---competing
concentrating concentrations of fragments that are significant on scale, which
then imply significance on a larger scale. 
This way, words that are clustered will tend to be more important than
words scattered liberally\footnote{In the past, I've used geometric
  decay of data in two-dimensional cylindrical
  timeseries\cite{burgessDSOM2002} to forget predictably.  Although
  cheap and mathematically `elegant', this works poorly in the regime
  of rapid sampling in a text document.  An exponential decay was too
  fast, so here a simple linear decay of memory potential was used.
  This gives an lesser impact on inverse importance sensitivity. By
  tuning it to label the most common words like `of' and `the' as
  least important.}.  The appeal of this method is that it's based on
realtime contextual competition between process rates, rather than
artificial sampling policies applied as a statistical afterthought.
Then the state learning is updating in real time, so we can't just
wait until the end of the document as treat the scanning as a
supervised training period, because by then the relative estimates
will have changed\footnote{As in the CFEngine case, the method is
  designed to run in unsupervised mode over the course of an input
  stream.}.

So, we approach the problem by treating each independent $n$-phrase fragment $\phi_n$
(according to some dissociation scheme) as a pseudo-periodic basis
vector that spans sentences and paragraphs within a document.

In a stream of words $w,w+1,etc$, the memory or learning level of a
given phrase at position $w_i$, $\phi_n(w_i)$, could be written $L_w(\phi_n)$. As new words arrive,
and $w\rightarrow w+1$, the memory is updated iteratively by the
partial function recursion rule on each new sample regardless
of whether $\phi_n$ is encountered or not: 
\beq 
L_{w+1}(\phi_n) = L_w(\phi_n) - \Delta L, 
\eeq 
for $L > \Delta L$, and further
\beq 
L_{w+1}(\phi_n) \rightarrow L_0
\eeq
each time $\phi_n$ is encountered anew, making $L_w \ge 0$. The linear decay
rate $\Delta L$ can be tuned to the recurrence frequency of the most
common words, so that a cognitive agent never forgets the common
words. Rather, it remembers them at a constant level and is thus
desensitized to them. On the other hand, more elaborate phrases are
more quickly forgotten so they come as a surprise again when then
recur later in a document. 

\begin{figure}[ht]
\begin{center}
\includegraphics[width=7cm]{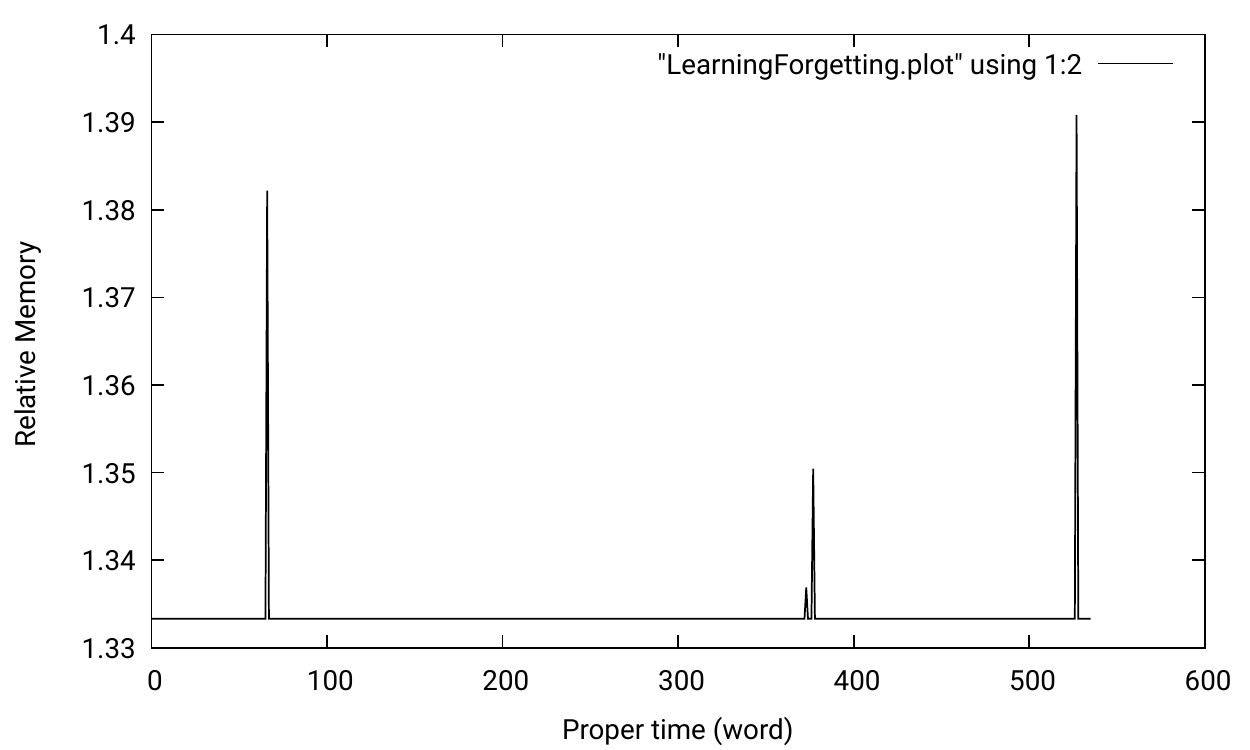}
\caption{\small As homogeneous words like `of' or `the' tick the
  learning clock, memory of frequent and homogeneously distributed
  words never has time to decay to zero, so their presence measures as
  ubiquitous. Apart from some small spikes, the level is quite steady
  with no apparent sawtooth pattern.\label{of_importance}}
\end{center}
\end{figure}

\begin{figure}[ht]
\begin{center}
\includegraphics[width=7cm]{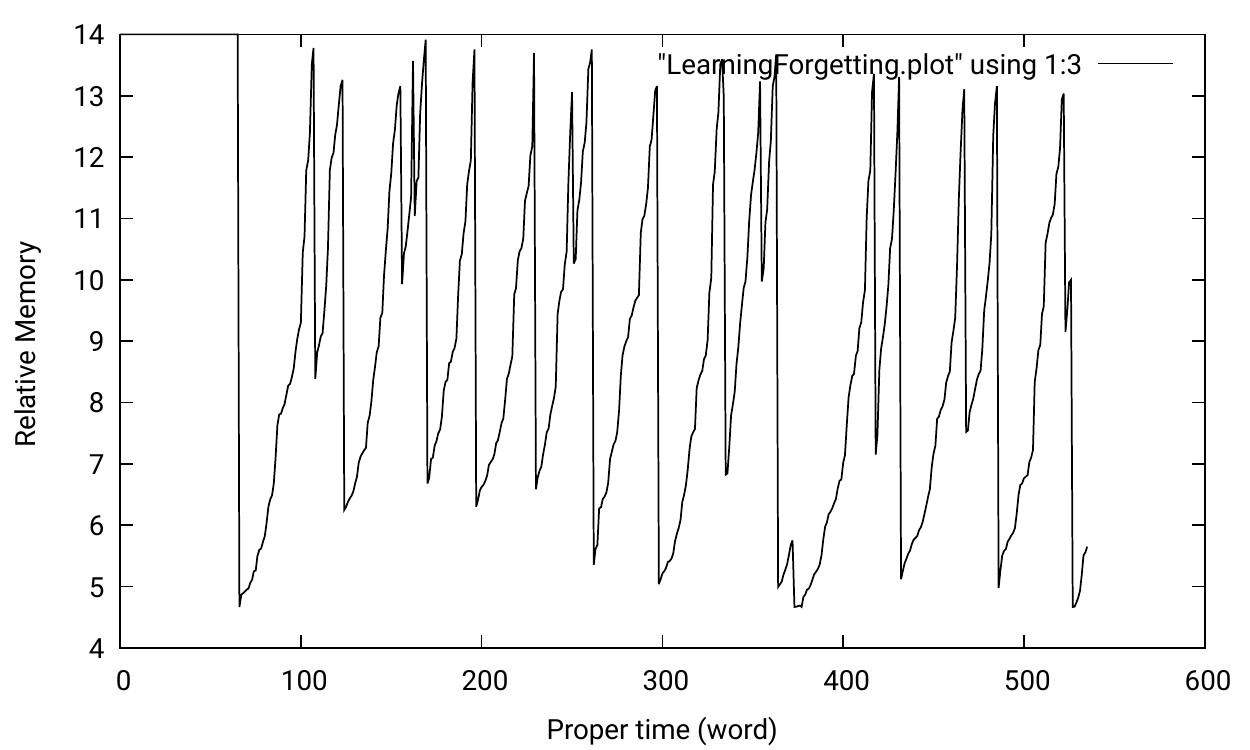}
\caption{\small Comparing to figure \ref{of_importance}, significant words like `economy' that a rarer and more
  inhomogeneously clustered than the covalent binding words (see spikes in the importance ranking), amplified by length and rarity, where the words
  occur, which decay eventually between
  cases.\label{economy_importance}}
\end{center}
\end{figure}

The unsupervised algorithm which approximates the measurement of
homogeneity therefore learns by identifying `surprises', whose effects
dissipate linearly with process time (so-called proper time by word
increment).  Surprises manifest as spikes in the regular quasi-periodic
process that forms the measurement scale (see figure \ref{of_importance} and \ref{economy_importance}
where we compare the word `of' with a word like `economy' in a certain
document).  Each new occurrence of a word is a spike.  For rare or
inhomogeneous words, the spike is easily contrasted against the
background for that basis function, at a certain scale $\Delta L$,
while such spikes smooth out at a more or less constant level for the
quasi-periodic, frequent and homogeneous words. In other words,
homogeneously distributed averages, over an entire narrative, become
platitudes--to which we desensitize quickly\footnote{This is probably
  how immune tolerance works too. In the immune system, there is
  probably a combination of evolutionary selectors and learned
  tolerance, an idea championed by the Danger Model. Looking for
  anomalous blips in a stream is a delicate and unstable matter, which
  is highly sensitive to scales. This is probably why organisms adapt
  to rather specific stable environments on evolutionary times, but
  fare worse over sudden variations.}.

Inhomogeneity in the stream (surprise) is the closest thing we can get
to a purely quantitative discriminator of significance. It explains
why physics typically abhors semantics: physics favours homogeneous
and isotropic phenomena, where simple steady-state rules may be
considered to govern time development. New information then enters a
steady-state system only at the boundaries of regions where the rules
are fixed.  The outcome of these experiments amounts to a weak claim
that we can explain sentence and language structure simply by
extracting proper names as multi-scale time-based patterns, and attaching them to
roles by composition and contextual networking.

\subsection{Discrimination of importance}

The threshold for attention activation is based on the number of words in a
fragment (not on the number of characters) and an arbitrary order of
magnitude for importance discrimination over each leg (say picking
base 10 arbitrarily). These choices seem to offer sampling rate
stability so that outcomes don't change spuriously for `fine tuning' of
parameters. If a result relies on fine tuning, then it isn't a real
phenomenon.

In the immunological approach, rival populations (relative
concentrations) of fragments (alphabetic molecules) compete for space
in memory. As one re-samples a bulk of changing populations, captured
as a memory buffer, fragments will persist until they are competed
away by others, so there will be natural hysteresis. The difference
between immunology and spacetime is how we treat an event: whether it
is important because many agents are talking about it
indistinguishably, or important because few are talking about distinctly.
On a microscopic level, this means that we are interested in patterns
that occur more than once over a `leg' of a narrative, but not too
often!  Meaning has a non-linear potential. The simplest way to handle
that is to distinguish newcomers by a lower level of significance than
repeated occurrences, even as all measures decay over time.

\subsection{Even selection and extraction}

The next step is to study whether or not we can extract a meaningful
subset of sentence `events' from streams, by way of capturing the key
concepts and summarizing their intended meaning, as judged by a human
arbiter. One would like to be able to talk about `topics', i.e. what
is being discussed in a sentence; however, that does not fall within
the scope of this method---as we recall that the Spacetime Hypothesis
has no comprehension of semantics, only local process scales.  The
ultimate goal is therefore not to summarize existing narratives but
rather to condense the data stream into chunks that could be associaed
with concepts. Eventually, this smart extraction might yield a `smart
sensor' which curates concepts more readily and places them relative
to one another in long-term memory. Then, a more sophisticated
cognitive process could generate narratives of its own, based in this
preprocessed summary. Moreover, this should work in different
processes too, such as in bioinformatics, system monitoring, diagnostics, etc.  This is
how the Spacetime Hypothesis presumes the emergence of automated
reasoning.

How then shall we decide what a stream is `about' without
semantics? What is the topic
of the monologue received by the sampling process?  If we reject prior
knowledge of words and their semantics, the measure and shape of the
monologue---its spacetime clustering properties---is the only
characteristic that remains as a discriminator.  This is also the only
basis for discriminating signals that a primitive organism would have
available to it, as it sought to evolve the capability of
communication. 

We attach significance to the features of the stream. A few thoughts come to
mind here:
\begin{itemize}
\item When telling stories, what we often note that the first remarks
  have specific importance, because listeners are more awake at the
  beginning, and because the story teller is trying to grab that
  attention.  This is a circular assumption, with some truth to it.
  It's probably a cultural norm. It could be turned into a more
  general principle that the sampler of information has a variable
  attentiveness at different stages of a document. Significant markers
  like beginning and end might be unnatural however. Not all
  narratives are told in a simple order.

\item Titles and headers are also assumed to carry meaning, when they
  function as a summary of a following passage. A title could be
  characterized as a sentence that stands alone in its own paragraph.
  However, a simple study of only a few cases shows that we must
  distrust titles that promise meaningful annotation. Not only is the
  identification of a `standalone paragraph' unreliable, but titles are not always
  representative of the content they lead.  Metaphors and artistic
  associations also label sections.  In a book about obsessions, a
  title might be called `The White Whale', referring to the cultural
  association with Moby Dick, but in fact be talking about dentistry.
  Some sections are numbered, but once again we have chosen to
  disregard special symbol meanings, with the exception of rudimentary
  punctuation.
\end{itemize}
As always, Promise Theory advises us to look for complementary processes (+ and - promises):
communication between autonomous unbiased agents must be a compromise
between positive offer of information (trusting a + promise) and 
negative selection (trusting a - promise).
Assessments of the listener are the ultimate selection arbiter.

\subsection{Realtime assessment by promise binding}

The implication of a promise theoretic binding is that meaning does
not arise solely from the source. Receivers of information from a
source pick and choose depending on their assessment of interest and
relevance to their own interior priorities.

An objective observer might try to honour the unaltered truth of the
source stream (only the + promises in a stream), while trying to keep
the - promises constant. However, in a sense that's just willfully
ignoring the obvious limitations and biases of the receiver.  A
Semantic Spacetime interpretation, on the other hand, predicts (from
its Promise Theory origins) that meaning is an assessment principally
of the receiver, conditioned by what remains of the source's
information once transmitted, filtered, wrecked by noise, and
received. The limitations of the receiver need to be
respected along with the receiver's process for assessment, for they
are what will determine the relevant observables in the
end\footnote{Technically, the parsing of text is harder than one might
  think. The highly variable representations of text, filled with
  special control characters and formatting instructions, make the
  identification of sentence and paragraph structure remarkably noisy.
  The technology for this will improve once the methodology has been
  understood.}.

Although we are not paying attention to the semantics of words, we
could explore the idea of words that make simple promises within a
stream (as noted in the introduction). Certain words act as catalysts for the formation of phrases.
These are the most commonly represented words too, from section on
bindings \ref{commonword}. There is a class of words that can't stand
on their own. The make very specific promises about ordering and
binding, so they could never be the start or the end of a meaningful
phrase. One could make an exclusion rule based on these promises. Such
a rule lies somewhere between a spacetime model (of causal order) and
an immunological model (of specific binding semantics). The results
of such a rule are of minor importance, however, as the ordinary
interferometry eventually serves the same function.

\subsection{Prototypical emotion and attention level}

Humans alter perceptions based on emotional state. Sudden changes in
emotion affect our attention levels and sampling rates of the
environment.  It seems likely that such dynamically changing
potentials play a useful role in regulating perception at all scales.
Whether one calls these changes emotions (semantics) or simply process
state biases (dynamics) is unimportant.

Human emotion clearly can be stimulated from the semantics of phrases,
building on what an agent experienced in the past, but the most basic
responses may be purely local, experiential, and dynamic in
nature---triggered by process discriminators that have been honed by
evolutionary timescales\cite{spacetime3}. These are the spacetime
discriminators that we're interested in here. These are what Kahneman
calls `system 1' discriminators: simple-minded blunt characterizations
that may be computed quickly for immediate sensory adaptation.

This idea is also not new.
In \cite{emotion2}, for instance, Jedud identifies `intensity' of language as well as
`interjection' (surprise factor) as emotional triggers. 
This idea can be incorporated as an aspect of spacetime process. Observers
respond to anomalous `surprise' events, bringing about increased heart
rate, and a heightened state of alertness.  It's an example of an
evolutionary scale adaptation, which must certainly impact the
interpretation of language too, by the principle of separation of
scales.  This increased interior time rate during a perceived crisis,
in turn, creates the relative perception that exterior time seems to
slow down during such a stressful episode\cite{observability}.
There is a large literature on looking for emotion
in text, in the sense of how humans identify the emotions expressed in
the document---using prior knowledge of parts of speech
and the meaning of emotion words (happy, sad, etc) to look for
co-activation clusters (see for instance \cite{emotion1,emotion2}) by
which the mood being expressed is described.  But this kind of
identification is not what we're looking for here.  Rather, we are
looking for the emotional bias of the reader of a text, and how that
feedback loop biases the sampling of the stream. When should we pay
special attention to a paragraph, and when can we browse past with a
minimum of attention?

We take such feedback mechanisms for granted, as humans, and expect to
pay more attention to passages where events are emotionally stressful
than for `boring' passages.  A simple way to experiment with this
contribution to meaning is to look for spacetime discriminators that
come as surprises and use them to increase the sampling rate of data
from the stream. Then we sample more often in a heightened state of
alertness, and allow that attentiveness to dissipate linearly (as with
the short term memory of input).  These landscape features then act as
simple signposts in the spacetime landscape---like anomalous spacing,
changes in average feature scale, punctuation marks.
Of the remaining observations, the relative importance may then be
used to select a representative subset.

\begin{figure}[t]
\begin{center}
\bigskip

\includegraphics[width=7.5cm]{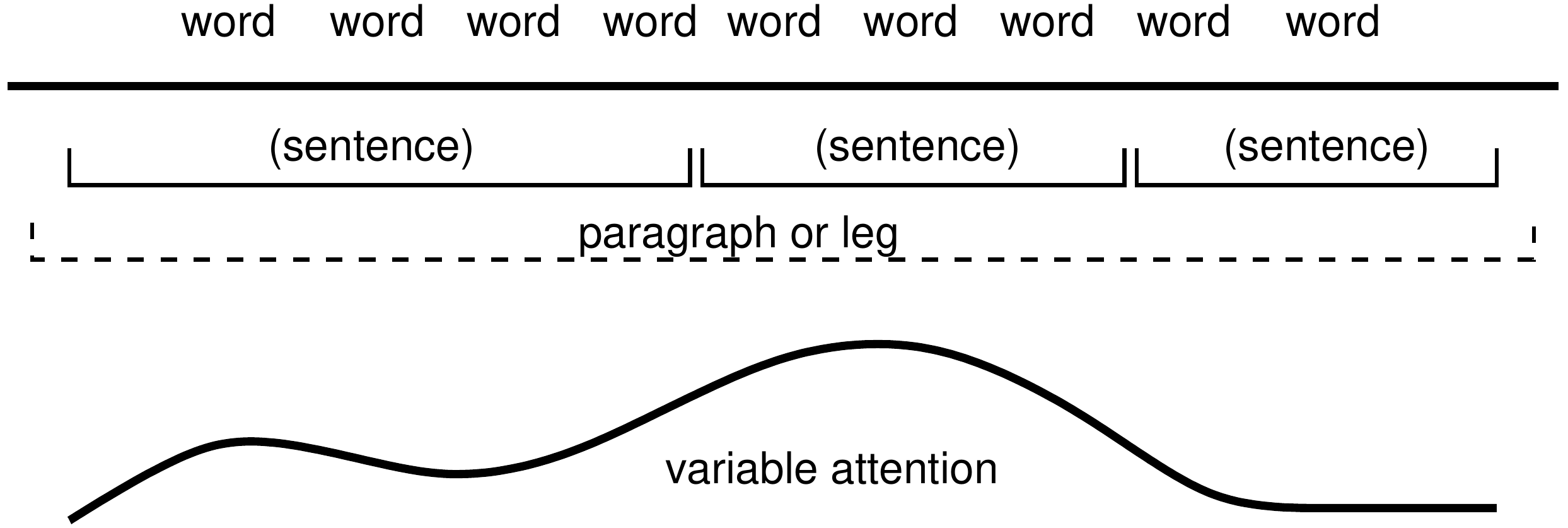}

\bigskip
\bigskip

\caption{\small Method for parsing text by sentence. Instead of paragraphs, encoded
variably in text, we look for natural `chunks' of text called `legs' of the
narrative counted by sentence.\label{text}}
\end{center}
\end{figure}

\begin{itemize}
\item Pruning by attention (see figure \ref{text}).
\item Pruning by importance (the meaning function).
\end{itemize}

The presence of special symbols, such as `!' might also indicate
stress in a generic way, while more fussy punctuation such as `;'
might be viewed as a reason to sigh and be sceptical of the length of
a sentence. However, this is already straying into the realm of symbol
semantics, which is to be avoided in this study.  Conversely,
interesting passages where we find important words that are repeated
might increase the sense that a narrative is `going somewhere',
whereas unresolved issues, in a stream of entirely new matters could
be stressful. Later, one could study the semantic triggers of emotion,
but for the scope of this paper we choose only to look only at the
spacetime measurables that are based on shifting scales.  Emotion
could be thought of as is a larger scale persistent state---and,
although a change could be triggered by a single word, it generally
persists for entire paragraphs, so it would become an attribute of the
way we assess context for each paragraph in a narrative.

There is a clear analogy here with the response of an immune system.
Certain events that trigger a response prime an immune system to move
into a certain state of operational awareness. Later, this dies down
and reverts to a passive state \cite{lisa98283}. However, if we're
honest in evaluatin the immunological approach, it was notable that
there was an almost complete failure to find sequence co-activations
within any of the documents.  Repeated co-activations were zero in all
cases, within the threshold of noise, which highlights the difference
between an immunological model and a spacetime model.

Narrative simply has too much intentional variation: there are no
repeated micro-episodes to speak of---so a statistical approach to
importance of specific symbols fails completely. The spacetime metric
approach, however, picks out 0.2\% of the document as standout
features, with a high degree of alignment with its supposed intent (by
human judgement). In this respect, the multi-scale Spacetime Hypothesis is the
clear winner between the two approaches.

\section{The evaluation of the text stream}

Based on the principles described above, we can now piece together a
realtime process for reading and ranking text, and summarize the
inputs using the assumption of focused narrative. Sentences are
dissociated into rolling fragments $\phi_n$, and these are counted
and permitted to `interfere' in the memory of the observer to
eliminate some and amplify others. Sentences are ranked according to
the importance of their fragments (up the the level of $6$-fragments).
A cutoff is applied to fragments that appear only once, and the importance
of fragments diminishes with usage. All this amounts to an importance ranking
which is not a static probability, but rather a dynamical process that involves
continuous learning {\em and forgetting} in balance.

\subsection{Arbitration of significant sentences}

The mixture of dynamical and semantic concepts in this study makes the
outcomes difficult to assess impartially---especially in quantitative
terms.  An example of the resulting output, produced by applying the
method to the President Obama Inaugural speech, is illustrated in
figure \ref{obama}. In order to decide whether the selected subsets
are representative of the original, we have to return to a kind of
Turing Test approach for evaluating them---by asking human test subjects for
their informal appraisal.

With available resources, this proved more challenging than expected.
Human test subjects bring their own ideas about purpose, interest,
expectations, and so on, to the appraisal, so they naturally look for
very different triggers when assessing the text. Humans read input
with far too much sophistication, unable to separate expectation from
reality, weaving their own ideas and expectations into sensory data,
and rejecting what doesn't fit their expectations.  Untrained humans
see what they want to see, not merely what is there.  That may be
obvious, on one level, but it certainly underlines the difference
between automated processes for data processing and reasoning and what
one might call Artificial Intelligence. For that reason, I'm inclined
to rate the success of the experiment as `not unsuccessful' and leave
it at that. The more interesting issue is whether the constructed narratives
in part 2 of the study make sense, and will be representative of the original
works.

None of the results point to the absolute supremacy of either an immunological
interpretation or a interpretation favouring the Spacetime Hypothesis,
because we---by using humans to judge the significance of the quoted
sentences as the selected events, we can't stop them from using the
semantics of the words as part of their judgement.  The best we can
hope for is to obtain a heuristic impression of whether or not:
\begin{itemize}
\item The intended relevant parts of the document were still perceived by the reader,
\item Any unintended interpretations of the text that could be made of the summary are no worse
than those that could be made of the original.
\end{itemize}
Comparing these is a tall order, but we must do what we can---as in the case of the Turing Test.
In short, the comments by readers were more aligned with my understanding of the texts
than I would have anticipated. However, the sample of readers was hardly representative of
the population at large, so it certainly warrants further testing.

\begin{figure*}
\begin{center}
\begin{quote}
\small
\begin{itemize}
\item \it We will not apologize for our way of life, nor will we waver in its defense, and for those who seek to advance their aims by inducing terror and slaughtering innocents, we say to you now that our spirit is stronger and cannot be broken.

\item What is required of us now is a new era of responsibility -- a recognition, on the part of every American, that we have duties to ourselves, our nation, and the world, duties that we do not grudgingly accept but rather seize gladly, firm in the knowledge that there is nothing so satisfying to the spirit, so defining of our character, than giving our all to a difficult task.

\item This is the meaning of our liberty and our creed -- why men and women and children of every race and every faith can join in celebration across this magnificent mall, and why a man whose father less than sixty years ago might not have been served at a local restaurant can now stand before you to take a most sacred oath.

\item Our  journey is not complete until our gay brothers and sisters are treated  like anyone else under the law –- applause -- for if we are truly  created equal, then surely the love we commit to one another must be  equal as well.

\item Our journey is not complete until we find a better way to  welcome the striving, hopeful immigrants who still see America as a  land of opportunity -- applause -- until bright young students and  engineers are enlisted in our workforce rather than expelled from our  country.

\item We must act, knowing that  todays victories will be only partial and that it will be up to those  who stand here in four years and years and years hence to  advance the timeless spirit once conferred to us in a spare  Philadelphia hall.++

\item \sc Sampled/Skipped =  153/123 of total 276, efficiency =  180.4 %
\end{itemize}
\end{quote}
\caption{\small Output from the sampling of President Obama Inaugural
  Speech taken as an example document of limited length that's easily
  relatable\cite{obama}. The selection of sentences is quite
  inhomogeneous, leaving a long gap at the start of the speech with
  preliminaries. This is interesting, given that one might expect
  transients at the start of a monologue to dominate.\label{obama}.}
\end{center}
\end{figure*}

General quantitative measurements of the summary can't serve as a
serious indicator of the success of the outcome.  One measure is the
efficiency in size reduction of a document. If the summary captures
essential parts of the narrative to an `acceptable level' then we can at least
measure how many words were considered redundant. The length of the original divided
by the length of the summary may be defined as the efficiency.

An arbitrary level of compression of a few hundred times was chosen to
test the idea at a non-trivial level. If too much of the original were
kept, one could not separate intended semantics from the outcome
plausibly. If too little were kept, it could not be representative.
The results sampled around 1-3 sentences per leg of 200 sentences
depending on variable attention.  Variations in attention level, from
the simple `emotional' response were deemed not distinguished enough
to warrant a wider range of variation. In practice, most legs got
around three important representatives and a few particularly
uninteresting ones were reduced to a single representative.
The limited emotional range represented by process anomalies didn't warrant
a greater variation than this.

Each text was of a different nature, with its own interesting
responses.  The case of the novel Slogans seems particularly
interesting, because a large novel could easily be a confusing
narrative, not really about a single topic. First, the high
compression rate for Slogans is likely due to the large number of
shorter conversational sentences present in the novel, which don't
drive the narrative forward in an obvious manner. In spacetime terms,
such short sentences are somewhat unremarkable unless they contain
very special phrases, and tend to be skipped.  The comparison of
summaries from the two different document formats offers a small
insight into how the summarization {\em doesn't} work. It's not simply
a matter of picking a constant rate of sentences that stand out.  It
cannot be related to the size of paragraphs, because the counting of
paragraphs by the algorithm differs by an order of magnitude, yet the
efficiency of summarization is almost identical in spite of the
parsing irregularity.  The ambiguities and complexities of the novel
were also captured by some of the readers in their comments. This
could be argued as a weakness, but the judgement seems accurate.  One
reader commented that it seemed like several stories in one, which is
indeed true.

\begin{table}[ht]
\begin{center}
\begin{tabular}{|c|c|l|}
\hline
Efficiency \% & Assess  & Name\\
\hline
 148 & related & History of Bede\\
 234 & related & Moby Dick\\
 129.5 & related & The Origin of Species (6th)\\
 207 & undecided & Random sentences\\
 497.5 & related & Slogans\\
 153 & undecided & Test 1 HTML\\
 196 & related & Test 2 HTML\\
 166 & related & Test 3 HTML\\
 171 & related & Test 4 HTML\\
 205 & related & Test 5 HTML\\
 180 & related & Obama speech\\
 152 & related & Fog and the Sea\\
\hline
\end{tabular}
\caption{\small Summarization of documents for a range of threshold
  values.  A compression of a few hundred times has been chosen to
  test the idea at a non-trivial level. This results in sampling
  around 1-3 sentences per leg of 200 sentences depending on variable
  attention.  The informal ranking of the extent to which the
  integrity of meaning survived the process is a heuristic impression
  based on feedback from a panel of 10 test subjects. This can be
  viewed as indicative, if not conclusive.\label{tab2}}
\end{center}
\end{table}
The coordinate alignment in parsing the two formats of Slogans is
slightly different, presumably due to stray formatting junk in the \LaTeX
version, which was not removed by preprocessing.  The method's crude
de-formatting saw a 5\% difference in the counting sentences between
HTML and \LaTeX versions of Slogans. 374 additional sentences, or a 7\%
difference in summary size.  Interestingly, running the method twice
on the same data did not see any changes in the selections, but
curiously the different formats did play a small role in changing the
selection of sentences.  In the Slogans case, it was possible to
consistently see the same sentences chosen from the respective
formats, with a large overlap between the two formats. However, there
were also differences in both samples.  Noise apparently causes text
differences (insertions like junk DNA) and small alterations in the
relative rankings, but once these are have passed over the memory
horizon, the summaries converge again.  Herein lies the importance of
forgetting data in a predictable and consistent manner---it
sets a timescale of its own, and allows
stability to be recovered even after deviations. This feels like a
reasonable explanation of the effect of noise.

Aside from compression rates, test subjects were asked to comment on
their subjective impressions of the summaries, without knowledge of
the original texts. This was difficult to test, and difficult to find
willing subjects to go along with the exercise, so the results are
perhaps broadly indicative at best.  The texts were sufficiently
unfamiliar to the audience that they could not have prior knowledge,
except in one instance.  From the summaries returned, I then made an
executive judgement on whether they had captured an essence of the
original. This turned out to work on a number of unanticipated
levels---not just about the presumed subject of the text, but on the
meta-level issues like organization and style, and indeed whether or
not they agreed with the sentiments being represented. Most readers
main emotional response to the text was based on this agreement or
disagreement with the apparent subject matter, which leans on their
advanced semantic understanding and thus doesn't play into the outcome
of this experiment per se.  

Since the outcomes have to be somewhat spurious, test subject
responses were only judged as representing either: related, unrelated,
or couldn't decide to assess whether the basic original intent was
captured or not.  Interestingly, only the random text was
`undecidable'. The actual intentional narratives did lead to an
intentional assessment.  At risk of reading too much into a simple
heuristic, I conclude that the use of spacetime structure to extract
the meaningful parts of a stream is `plausible', and would encourage
others with greater resources to pursue the idea.

Finally, with a simple understanding of how quantitative measures
behave within streams of symbolic data, alongside the ability to
extract key events (ranked to be meaningful by the extent to which
they stand out in the information landscape), one can now go on to
construct a relational model for the events, as proposed in
\cite{cognitive}.  That result will be presented in the sequel to this
paper.

\section{Summary and Remarks} 

The goal of this pre-study was to examine whether spacetime
structures, within sensory data, might be sufficient to bootstrap the
identification meaningful concepts in a process of unsupervised
learning.  The results suggest that what we consider important and
interesting about sensory experience need not be based solely on
higher reasoning, but that simple spacetime process cues suffice to
bootstrap such cognition at the outset.  A rival hypothesis about
signal co-activation, in the manner of an Artificial Immune System,
was not able to develop criteria to select meaningful sentences.
However, the interferometric method used here could be considered a hybrid between an
Artificial Immune System co-activation model, and spacetime
discriminator.  Spacetime scales arise from short-range ordered
symbols, analogous to features in matter or vacuum state in physics.
Symbols are not strongly ordered in narrative, so there is 
long-range order.

The study lends weight to the Spacetime Hypothesis for knowledge
seeding.  Meaningful signals can indeed be extracted, with a high
degree of consistency, from a narrative stream, and are resilient to
small variations in scale.  The method is not so stable as to be
completely unaffected by changes---which is good---but it is stable
enough that concepts retain their invariant meanings, in spite of
considerable surrounding noise. 

From this first part of the study, one develops guidelines for a
specially adapted `smart sensor', capable of preprocessing a raw data
stream in order to mine it for the seeds of concepts. From here, one
expects to build on the results and the semantics of the spacetime
model to enable narrative reconstruction\cite{cognitive}.  

The spacetime hypothesis offers and interesting insight into the way
we receive narrative information in samples. It might seem like an odd
idea to see how little we can get away with---without understanding
concepts like grammar and parts of speech. Pre-liguistic (Information
Theoretic) considerations suffice to large extent.  This might also
weigh in on Chomsky's controversial idea of the level of innateness in
linguistic capability\footnote{Compare this work to the study in
  \cite{chomgrammar} for example.}, and on generative grammars in
general\cite{pinker1,unfolding,langacker1}.
Given the limitations of this study, there's clearly a lot more depth
one could plumb when it comes to finding principles behind a smart
sensor. 

What stands out in this work is the role of scales and their
interaction. This supports the Promise Theory model and its separation
of scales as superagents\cite{promisebook,spacetime2}.  The Spacetime
Hypothesis posits that the semantics of text are ultimately derivative
of the spacetime structures and scales in sensory input. Over long
times and many episodes, one imagines that patterns will compound
together to form increasingly sophisticated distinctions---and
therefore concepts.  Negative selection plays a small role here
already (in attentiveness pruning, for instance), and one would expect
it to play further roles in post processing too. This will come to the
fore in the second part of the study.

The deliberate restriction to small amounts of contextualized sensory
data, in unsupervised mode, is key to this study.  Active agents
(those in the field, so to speak), cannot rely on large curated data
sets, except for what's implicit in their genetic makeups. Big data
studies may impress us with the significance of large numbers, but
there are substantial and well-understood weaknesses to relying on
large $N$: it wipes out a lot of structural information.

In the final instance, we would like to apply the technique developed
here to other sources of data that are less well understood---such as
data from mechanical and information systems, where complexity masks
the causal chain between semantics and dynamics. It remains to be seen
what we could infer from other kinds of timeseries data, using purely
quantitative numerical alphabets. Most likely, the effective pattern
set leads to a much smaller vocabulary, so several independent signals
may have to be combined to infer sufficient discriminatory context and
establish a meaning.  An interferometry based on Fourier components
might fit the analogue signal model better.

As with many papers that fall under the umbrella of Artificial
Intelligence, this work ultimately has the flavour of `an interesting
magic trick', which possibly points to some underlying science yet to
be fully understood.  What seems nice about the result is the way it
seems possible to rank meaning and thence intent based purely on
metric differentiation. In the bigger picture, this is the very
foundation of the Newtonian approach to mechanics, here applied to
a discrete symbolic spacetime.  That should be a wakeup call to those
argue for artificial distinctions between `symbolic AI' and `deep
learning' methods. If one understands scales properly, and treats
changes impartially, the distinction can only be about scales.

{\sc Acknowledgement:} I'm thankful to those who gave their time as test
subjects and offered comments on the work.

\bibliographystyle{unsrt}
\bibliography{spacetime}

\end{document}